\documentclass[10pt,journal,compsoc]{IEEEtran}
\usepackage{cite}

\usepackage{graphicx}
\usepackage{amsmath}
\usepackage{amssymb} 
\usepackage{array}
\usepackage{url}
\usepackage{color}
\usepackage{booktabs}
\usepackage{subfloat}
\usepackage{float}
\usepackage{subfig}
\usepackage{bbding}
\usepackage{textcomp}
\usepackage{multirow}
\usepackage{overpic}
\usepackage{amsfonts}
\usepackage{xcolor,colortbl}

\usepackage{algorithm}
\usepackage{algorithmicx}
\usepackage{algpseudocode}

\definecolor{Gray}{gray}{.90}

\newcommand{\tabincell}[2]{\begin{tabular}{@{}#1@{}}#2\end{tabular}}

\newcommand{\etal}{\textit{et al}.}
\newcommand{\ie}{\textit{i}.\textit{e}.}
\newcommand{\eg}{\textit{e}.\textit{g}.}
\newcommand{\etc}{\textit{etc}}

\newcommand{\changes}{}

\ifCLASSINFOpdf

\else

\fi

\usepackage[bookmarks,colorlinks,linkcolor=red, anchorcolor=blue, citecolor=green]{hyperref}
\begin{document}

\title{Self-Supervised Learning for Real-World Super-Resolution from Dual and Multiple Zoomed Observations}
\author{Zhilu Zhang, Ruohao Wang, Hongzhi Zhang, and Wangmeng Zuo,~\IEEEmembership{Senior Member,~IEEE}
\IEEEcompsocitemizethanks{\IEEEcompsocthanksitem Z. Zhang is with the Faculty of Computing, Harbin Institute of Technology, Harbin, China. (E-mail: cszlzhang@outlook.com)
\IEEEcompsocthanksitem R. Wang is with the Faculty of Computing, Harbin Institute of Technology, Harbin, China. (E-mail: rhwangHIT@outlook.com)
\IEEEcompsocthanksitem H. Zhang is with the Faculty of Computing, Harbin Institute of Technology, Harbin, China.  (Corresponding author. E-mail: zhanghz0451@gmail.com)
\IEEEcompsocthanksitem W. Zuo is with the Faculty of Computing, Harbin Institute of Technology, Harbin, China. (E-mail: wmzuo@hit.edu.cn)
}
}

\markboth{Journal of \LaTeX\ Class Files,~Vol.~14, No.~8, August~2015}%
{Shell \MakeLowercase{\textit{et al.}}: Bare Demo of IEEEtran.cls for Computer Society Journals}

\IEEEtitleabstractindextext{%
\begin{abstract}
  In this paper, we consider two challenging issues in reference-based super-resolution (RefSR) for smartphone, (i) how to choose a proper reference image, and (ii) how to learn RefSR in a self-supervised manner. 
  Particularly, we propose a novel self-supervised learning approach for real-world RefSR from observations at dual and multiple camera zooms.
  Firstly, considering the popularity of multiple cameras in modern smartphones, the more zoomed (telephoto) image can be naturally leveraged as the reference to guide the super-resolution (SR) of the lesser zoomed (ultra-wide) image, which gives us a chance to learn a deep network that performs SR from the dual zoomed observations (DZSR).
  Secondly, for self-supervised learning of DZSR, we take the telephoto image instead of an additional high-resolution image as the supervision information, and select a center patch from it as the reference to super-resolve the corresponding ultra-wide image patch. 
  To mitigate the effect of the misalignment between ultra-wide low-resolution (LR) patch and telephoto ground-truth (GT) image during training, we first adopt patch-based optical flow alignment to obtain the warped LR, then further design an auxiliary-LR to guide the deforming of the warped LR features.
  To generate visually pleasing results, we present local overlapped sliced Wasserstein loss to better represent the perceptual difference between GT and output in the feature space.
  During testing, DZSR can be directly deployed to super-solve the whole ultra-wide image with the reference of the telephoto image.
  In addition, we further take multiple zoomed observations to explore self-supervised RefSR, and present a progressive fusion scheme for the effective utilization of reference images.
  Experiments show that our methods achieve better quantitative and qualitative performance against state-of-the-arts.
  Codes are available at \url{https://github.com/cszhilu1998/SelfDZSR_PlusPlus}.
\end{abstract}

}

\maketitle

\IEEEdisplaynontitleabstractindextext

\IEEEpeerreviewmaketitle

\begin{figure}[t]
	\centering
	\begin{overpic}
	  [width=0.99\linewidth]{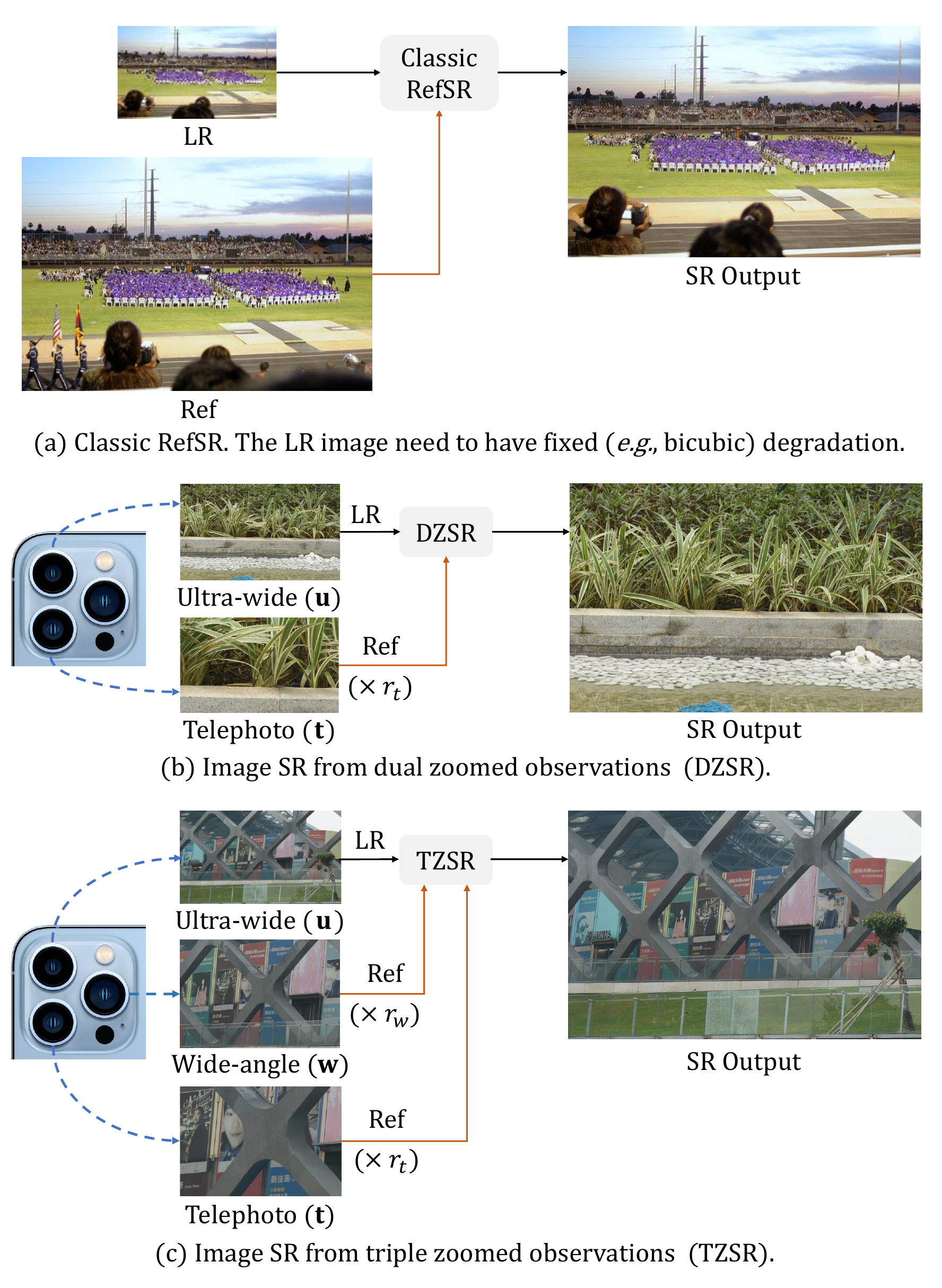}
	\end{overpic}
        \vspace{-2mm}
	\caption{Overall pipeline of the classic RefSR, DZSR, and TZSR during inference. $r_w$ and $r_t$ respectively represent the wide-angle and telephoto resolution multiple relative to the ultra-wide image.}
	\label{fig:intro}
  \vspace{-3mm}
  \end{figure}

\IEEEraisesectionheading{\section{Introduction}\label{sec:introduction}}
  \IEEEPARstart{I}{mage} super-resolution (SR)~\cite{SRCNN,SRGAN,EDSR,RCAN,Swinir} aiming to recover a high-resolution (HR) image from its low-resolution (LR) counterpart is a severely ill-posed inverse problem with many practical applications. 
  Recently, reference-based image SR (RefSR)~\cite{SRNTT,FRM,SSEN,TTSR,MASA-SR,C2-Matching,huang2022task,cao2022reference,zhang2022rrsr,DCSR} has made progress in relaxing the ill-posedness, which suggests to super-resolve the LR image for more accurate details by leveraging a reference (Ref) image, as shown in Fig.~\ref{fig:intro}(a).
  
  For RefSR, the Ref image should contain similar content and texture with the HR image, and is generally acquired from video frames (\eg, CUFED5 dataset~\cite{SRNTT}) and web image search (\eg, WR-SR dataset~\cite{C2-Matching}).
  However, the video frame with high resolution cannot be always got in realistic scenarios, while web image retrieval is time-consuming and sometimes unreliable.
  It remains a challenging issue to choose a proper Ref image for each LR image, especially in real-world applications. 
  Fortunately, advances and popularity of imaging techniques make it practically feasible to collect images of a scene at different camera zooms. 
  For example, asymmetric cameras with different fixed-focal lenses have been equipped in modern smartphones.
  In these practical scenarios, the more zoomed (telephoto) image can be naturally leveraged as the reference to guide the SR of the lesser zoomed (ultra-wide) image. 
  Image SR from the dual zoomed observations (DZSR) can thus be carried out, in which Ref has the same scene as the center part of the LR image but higher resolution, as shown in Fig.~\ref{fig:intro}(b). 

  DZSR is different from classic RefSR methods, but can be still regarded as a special case of RefSR.
  While conventional RefSR methods~\cite{SRNTT,FRM,SSEN,TTSR,MASA-SR,C2-Matching,huang2022task,cao2022reference,zhang2022rrsr} usually use synthetic (\eg, bicubic) degraded LR images for training and evaluation, DZSR should cope with real-world LR ultra-wide images and no ground-truth HR images are available in training. 
  To bridge the domain gap between synthetic and real-world LR images, DCSR~\cite{DCSR} suggests a self-supervised real-image adaptation (SRA) strategy, which involves degradation preserving and detail transfer terms. 
  However, DCSR~\cite{DCSR} only attains limited success, since the two loss terms in SRA cannot well address the gap between the synthetic and real-world LR degradation as well as the misalignment between ultra-wide and telephoto images.
  Different from DCSR~\cite{DCSR} requiring to pre-train on synthetic images, we introduce self-supervised learning to train DZSR model from scratch directly on ultra-wide and telephoto images, without additional HR images as ground truths (GT).
  Specifically, we crop the center part of the ultra-wide and telephoto images respectively as the input LR and Ref images, and use the whole telephoto image as the GT during training (see Fig.~\ref{fig:SelfDZSR}(a)). 
  During inference, by taking the whole ultra-wide and telephoto images respectively as LR and Ref, DZSR can be directly deployed to super-solve the whole ultra-wide image (see Fig.~\ref{fig:SelfDZSR}(c)).
  
  However, when training DZSR model, the cropped ultra-wide LR image generally cannot be accurately aligned with the telephoto GT image, making the learned model prone to producing blurry SR results~\cite{RAW-to-sRGB,SRRAW}.  
  In this case, matching the Ref to LR will also result in the warped Ref being not aligned with the GT, bringing more uncertainty to network training.
  To handle the misalignment issue, we propose a two-stage alignment method.
  Firstly, we perform patch-based alignment between LR and GT images by a pre-trained optical flow network.
  The warped LR is thus obtained, which is roughly aligned with GT. 
  Secondly, we perform finer alignment in the feature space.
  Specifically, we hope to construct an auxiliary-LR as the target position image for deforming the warped LR features toward GT during training.
  The auxiliary-LR should be aligned with GT and can be replaced by LR safely during inference.
  Thus, we carefully design the auxiliary-LR generator network, as well as its position preserving and content preserving constraints.
  Then an adaptive spatial transformer network (AdaSTN) is used to deform the warped LR features for obtaining final aligned LR features, according to offsets estimated between the warped LR and auxiliary-LR features.
  When training is done, the optical flow network, auxiliary-LR generator and offset estimator of AdaSTN can be safely detached, bringing no extra cost in the test phase.

  For the matching of the Ref image, we perform corresponding contents searching similar to most existing RefSR methods~\cite{SRNTT,TTSR,MASA-SR,C2-Matching,DCSR}, but it is from Ref to the warped LR image rather than from Ref to the original LR image.
  Finally, the aligned LR and aligned Ref features can be combined and fed into the restoration module.
  Furthermore, we present local overlapped sliced Wasserstein (LOSW) loss to optimize the DZSR model.
  LOSW loss can better measure the perceptual difference between GT and output in the feature space, which is beneficial to the generation of visually pleasing results.

  In addition, we bring self-supervised RefSR from multiple zoomed observations into account, and take SR from triple zoomed ones (TZSR) as an example to achieve.
  For TZSR, the wide-angle image can be utilized as an addition Ref, which resolution is between the resolution of ultra-wide and telephoto images, as shown in Fig~\ref{fig:intro}(c).
  Following the self-supervised learning of DZSR, self-supervised TZSR can be succeeded.
  Moreover, to make better use of reference images, we present a progressive fusion scheme for TZSR, where the aligned Ref features (from wide-angle and telephoto images) successively fuse with the aligned LR ones.

  Extensive experiments are conducted on the Nikon camera images from the DRealSR dataset~\cite{CDC} as well as the iPhone camera images from the RefVSR dataset~\cite{RefVSR}. 
  The results demonstrate the effectiveness and practicability of the proposed method for real-world RefSR.
  In comparison to the state-of-the-art SR and RefSR methods, our method performs favorably in terms of both quantitative metrics and perceptual quality. 
  We also conduct detailed ablation studies, analyzing the effectiveness of different components in the proposed method.

  In comparison with the previous version SelfDZSR~\cite{SelfDZSR} in ECCV 2022, two main changes (patch-based optical flow alignment and LOSW loss) are introduced to improve the self-supervised learning pipeline for DZSR, while TZSR is newly proposed in this work.
  The proposed self-supervised learning framework for DZSR and TZSR is named SelfDZSR++ and SelfTZSR++, respectively.
  To sum up, the main contributions of this work include:
  \begin{itemize}
    \item To achieve real-world RefSR from dual zoomed observations without additional HR images, we propose a self-supervised framework SelfDZSR++.
    \item To alleviate the adverse effect of image misalignment for self-supervised learning, we propose a two-stage alignment method involving patch-based optical flow alignment and auxiliary-LR guiding alignment, while bringing no extra inference cost. %
    \item To generate visually pleasing results, we present local overlapped sliced Wasserstein loss for measuring perceptual differences better. %
    \item To explore self-supervised RefSR from multiple zoomed observations, SelfDZSR++ is expanded to SelfTZSR++, where we present a progressive fusion scheme for efficient restoration. %
    \item Quantitative and qualitative results on the Nikon and iPhone camera images show that our method outperforms the state-of-the-art methods. 
  \end{itemize}

\section{Related Work} 
\subsection{Blind Single-Image Super-Resolution}
 With the development of deep networks, single image super-resolution (SISR) methods based on fixed and known degradation have achieved great success in terms of both performance~\cite{SRCNN,SRGAN,EDSR,RCAN,SRMD,Swinir} and efficiency~\cite{IMDN,AdaDSR,wang2021exploring,ClassSR,Xie_2021_ICCV}.
  However, these methods perform poorly when applied to images with unknown degradations, and may cause some artifacts.
  Thus, blind super-resolution comes into being to bridge the gap.
  
  On the one hand, some works estimate the blur kernel or degradation representation for LR and feed it into the SR reconstruction network.
  IKC~\cite{IKC} performs kernel estimation and SR reconstruction processes iteratively, while DAN~\cite{DAN} conducts it in an alternating optimization scheme.
  KernelGAN~\cite{KerGAN} utilizes the image patch recurrence property to estimate an image-specific kernel, and FKP~\cite{FKP} learns a kernel prior based on normalization flow~\cite{flow} at test time.
  To relax the assumption that blur kernels are spatially invariant, MANet~\cite{MANet} estimates a spatially variant kernel by the suggested mutual affine convolution.
  Different from the above explicit methods of estimating kernel, DASR~\cite{DASR} introduces contrastive learning~\cite{he2020momentum} to extract discriminative representations to distinguish different degradations. 
  On the other hand, Hussein~\etal~\cite{correction_cvpr2020} modify the LR to a pre-defined degradation type (\eg, bicubic) by a closed-form correction filter.
  BSRGAN~\cite{BSRGAN} and Real-ESRGAN~\cite{Real-ESRGAN} design more complex degradation models to generate LR data for training the networks, making the networks generalize well to many scenarios with real-world degradation.
  %

\subsection{Real-World Single-Image Super-Resolution} 
  Although blind SR models trained on synthetic data have shown appreciable generalization capacity, the formulated degradation assumption limits the performance on real-world images with much more complicated and changeable degradation.
  Thus, image SR directly toward real-world scenes has also received much attention.
  On the one hand, given unpaired real LR and HR, several real-world SR methods~\cite{NTIRE2020,AIM2019,wei2021unsupervised} attempt to approximate real degradation and generate the auxiliary-LR image from HR. Then they learn to super-resolve the auxiliary-LR in a supervised manner.
  \changes{On the other hand, some methods~\cite{CameraLen,LP-KPN,CDC,AIM2020,NTIRE2019} construct paired datasets by adjusting the focal length of a camera, in which the image with a long focal and short focal length is regarded as GT and LR, respectively.
  %
  %
  In this work, our data collection manner is similar to theirs.
  The main difference lies in the tasks that need to be performed.
  We entail training a RefSR model rather than a SISR model, where images with different focal lengths are all required to be taken as input.}

  In addition, spatial misalignment is a universal problem in real-world paired datasets, and it may cause blurry SR results.
  The above methods based on paired datasets pre-execute complex alignment or even manual selection, which are generally laborious and time-consuming.
  Different from them, CoBi~\cite{SRRAW} loss offers an effective way to deal with misalignment during SR training.
  Zhang~\etal~\cite{RAW-to-sRGB} incorporates global color mapping and optical flow~\cite{PWC-Net} to explicitly align the data pairs with severe color inconsistency.
  Nevertheless, optical flow is limited in handling complicated misalignment.
  In this work, we further propose patch-based optical flow alignment and auxiliary-LR guiding deforming to handle the complicated misalignment after image-based alignment with optical flow.

\subsection{Reference-Based Image Super-Resolution}
  RefSR aims to take advantage of a high-resolution reference image that has similar content and texture as HR image for super-resolution.
  It relaxes the ill-posedness of SISR and facilitates the generation of more accurate details.
  The features extracting and matching between LR and Ref is the research focus of most RefSR methods.
  Among them, Zheng~\etal~\cite{refsr_bwvc} proposes a correspondence network to extract features for matching, and an HR synthesis network with the input of the matched Ref.
  SRNTT~\cite{SRNTT} calculates the correlation between pre-trained VGG features of LR and Ref at multiple levels for matching them.
  Zhang~\etal~\cite{zhang2020texture} extend the scaling factor of RefSR methods from $4\times$ to $16\times$.
  Furthermore, TTSR~\cite{TTSR} and FRM~\cite{FRM} develop an end-to-end training framework and proposed learnable feature extractors.
  $C^2$-Matching~\cite{C2-Matching} performs a more accurate match by the teacher-student correlation distillation.
  MASA-SR~\cite{MASA-SR} reduces the computational cost by coarse-to-fine correspondence matching.
  Recently, Huang~\etal~\cite{huang2022task} decouples the RefSR task into SISR and the texture transfer tasks for alleviating reference-underuse and reference-misuse issues, while RRSR~\cite{zhang2022rrsr} introduces the reciprocal learning strategy to improve RefSR models.
  Besides, CrossNet~\cite{CrossNet} and SEN~\cite{SSEN} respectively introduce optical flow~\cite{FlowNet} and deformable convolution~\cite{DefConv,DefConv-v2} to align Ref with LR.
  However, optical flow is limited in handling large and complicated motions while deformable convolution is limited in modeling long-distance correspondence.
  In this work, we follow~\cite{C2-Matching} to perform patch-wise matching.

  Additionally, the RefSR methods mentioned above are all based on bicubic down-sampling.
  DCSR~\cite{DCSR} explores an adaptive fine-tuning strategy on real-world images based on the pre-trained model with synthetic data.
  In this work, we propose a fully self-supervised learning framework directly on weakly aligned multiple zoomed observations.

  \begin{figure*}[t]
	\centering
	\begin{overpic}
	  [width=0.98\linewidth]{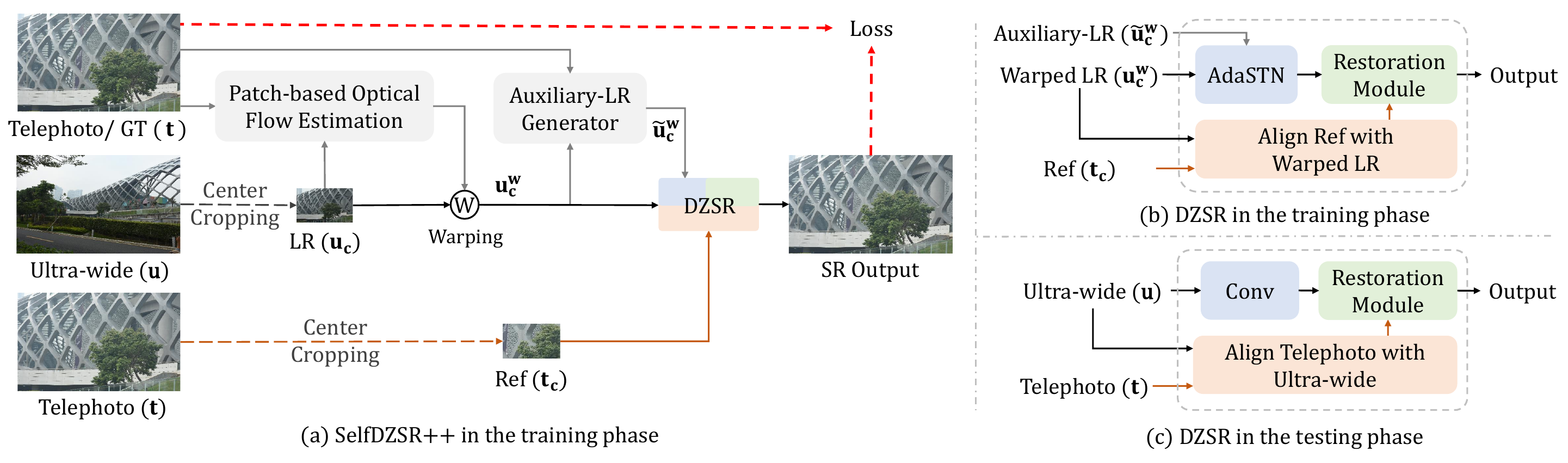}
	\end{overpic}
        \vspace{-2mm}
	\caption{Overall pipeline of proposed SelfDZSR++. (a) SelfDZSR++ in the training phase. The original telephoto image is taken as GT ($\mathbf{t}$), while the center areas of ultra-wide ($\mathbf{u}$) and telephoto ($\mathbf{t}$) images are regarded as LR ($\mathbf{u_c}$) and Ref ($\mathbf{t_c}$)  images, respectively. (b)~DZSR in the training phase. The auxiliary-LR ($\mathbf{\tilde{u}_c^w}$) is aligned with GT and used for deforming the warped LR ($\mathbf{u_c^w}$) towards the GT by AdaSTN. Then aligned LR and Ref features are fed into the restoration module. (c)~DZSR in the testing phase. The ultra-wide ($\mathbf{u}$) and telephoto ($\mathbf{t}$) images can be regarded as LR and Ref, respectively.
    Patch-based optical flow alignment and auxiliary-LR generator are detached. AdaSTN is simplified to a convolution layer.}
	\label{fig:SelfDZSR}
  \vspace{-2mm}
  \end{figure*}

\section{Proposed Method} 
  In this section, we first introduce our self-supervised learning approach of SelfDZSR++. 
  Then we detail the solutions for alignment between LR and GT, alignment between Ref and LR, restoration module, LOSW loss, and learning objective in SelfDZSR++.
  Finally, we propose an extension of SelfDZSR++, which utilizes multiple zoomed observations to perform self-supervised RefSR.
  %

\subsection{Self-Supervised Learning Framework} \label{sec:self-supervised}
  %
  Denote by $\mathbf{u}$ and $\mathbf{t}$ the ultra-wide image and the telephoto image, respectively.
  Super-resolution based on dual zoomed observations aims to super-resolve the ultra-wide image $\mathbf{u}$ with the reference telephoto image $\mathbf{t}$, which can be written as, 
    \begin{equation}
      \hat{\mathbf{y}} = \mathcal{Z}(\mathbf{u}, \mathbf{t}; \Theta_\mathcal{Z}),
    \label{eqn:DZSR}
    \end{equation}
  where $\hat{\mathbf{y}}$ has the same field-of-view as $\mathbf{u}$ and the same resolution as $\mathbf{t}$, $\mathcal{Z}$ denotes the zooming network with the parameter $\Theta_\mathcal{Z}$.
  
  However, in real-world scenarios, the GT of $\hat{\mathbf{y}}$ is hard or almost impossible to acquire.
  A simple alternative solution is to leverage synthetic data for training, but the domain gaps between the degradation model of synthetic images and that of real-world images prevent it from working well.
  DCSR~\cite{DCSR} tries to bridge the gaps by fine-tuning the trained model using an SRA strategy, but the huge difference in the field of view between the output and the target telephoto images limits it in achieving satisfying results.
  
  In contrast to the above methods, we propose a novel self-supervised dual-zooms super-resolution (SelfDZSR++) framework, which can be trained from scratch solely on the ultra-wide and telephoto image (see Fig.~\ref{fig:SelfDZSR}(a)), and be directly deployed to the real-world dual zoomed observations (see Fig.~\ref{fig:SelfDZSR}(c)).
  During training, we first crop the central area of the ultra-wide and telephoto images,
    \begin{equation}
      \mathbf{u_c} = \mathcal{C}(\mathbf{u}; r_t), \qquad
      \mathbf{t_c} = \mathcal{C}(\mathbf{t}; r_t),
    \label{eqn:crop}
    \end{equation}
  where $\mathcal{C}$ denotes the center cropping operator, $r_t$ is the focal length ratio between $\mathbf{t}$ and $\mathbf{u}$.
  Note that $\mathbf{t_c}$ has the same scene and higher resolution with $\mathcal{C}(\mathbf{u_c}; r_t)$, \ie, the central area of $\mathbf{u_c}$.
  Simultaneously, the resolution of $\mathbf{t}$ is $r_t$ times that of $\mathbf{u_c}$, and their scene is the same.
  Thus, $\mathbf{u_c}$ and $\mathbf{t}$ can be naturally used as LR and GT respectively, while $\mathbf{t_c}$ can be regarded as the Ref during training.
  Then we can define DZSR as,
    \vspace{-1.5mm}  
    \begin{equation}
      \Theta_\mathcal{Z}^{*} = \arg \min_{\Theta_\mathcal{Z}} \mathcal{L}\left(\mathcal{Z}(\mathbf{u_c}, \mathbf{t_c}; \Theta_\mathcal{Z}), \mathbf{t} \right),
    \label{eqn:SelfDZSR}
    \end{equation}
  where $\mathcal{L}$ denotes the self-supervised learning objective.

  Nonetheless, GT $\mathbf{t}$ is not spatially aligned with LR $\mathbf{u_c}$, bringing adverse effects on self-supervised learning.
  To handle the misalignment issue, we want to align LR to GT as much as possible during training.
  And we hope such an operation won't affect the inference process.
  For this purpose, the elaborate design of the framework is essential for SelfDZSR++, which is introduced below.

\subsection{Alignment between LR and GT} \label{sec:alignment_lr}
For aligning LR to GT, we propose a two-stage alignment method.
First, we adopt patch-based optical flow alignment to get a warped LR image that is roughly aligned with GT.
Then we construct an auxiliary-LR to guide the deformation of warped LR towards GT in the feature space, which is more refined. 
\subsubsection{Patch-based Optical Flow Alignment}
  The LR image $\mathbf{u_c}$ and GT image $\mathbf{t}$  in SelfDZSR++ are captured from the different camera lenses, and are generally misaligned in space.  
  When training the model using these pairs, the output would be spatially misaligned with GT, thus leading to inaccurate pixel-wise loss calculation.
  And it has been shown in recent works~\cite{SRRAW,RAW-to-sRGB} that such misalignment will cause the network to generate blurry results.
  More seriously, the misalignment will result in the warped Ref features being not aligned with GT after matching Ref to LR, bringing more uncertainty to model learning. 
  
  Off-the-shelf optical flow~\cite{PWC-Net} offers a probable solution to deal with this issue.
  But when the image resolution is large (\eg, $>$1K), the optical flow network sometimes tends to estimate the motion globally and performs poorly on small local contents.
  Thus, we further adopt patch-based optical flow alignment after image-level alignment, which is carried out on the training patches cropped from the original images.
  Specifically, we take PWC-Net~\cite{PWC-Net} to calculate the optical flow from the GT patch to the LR one.
  Then we back warp the LR patch to get the warped LR $\mathbf{u_c^w}$ according to the optical flow.

\subsubsection{Generation of Auxiliary-LR for Alignment} 
  However, limited to the offset diversity~\cite{chan2021understanding} of optical flow, the warped LR $\mathbf{u_c^w}$ and GT $\mathbf{t}$ are still slightly misaligned in some complex circumstances (\eg, occlusions caused by scene parallax or moving objects) and explicit perfect alignment is impracticable. 
  To handle the above issues, we hope to construct an auxiliary-LR $\mathbf{\tilde{u}_c^w}$ from the GT $\mathbf{t}$ while keeping the spatial position unchanged, and take it to guide the alignment of warped LR towards GT in the feature space (see Fig.~\ref{fig:SelfDZSR}(a) and (b)).
  Noted that the auxiliary-LR cannot be used in testing, and it should be substituted by the ultra-wide $\mathbf{u}$ (see Fig.~\ref{fig:SelfDZSR}(c)).
  
  \begin{figure}[t]
    \centering
    \begin{overpic} 
      [width=0.98\linewidth]{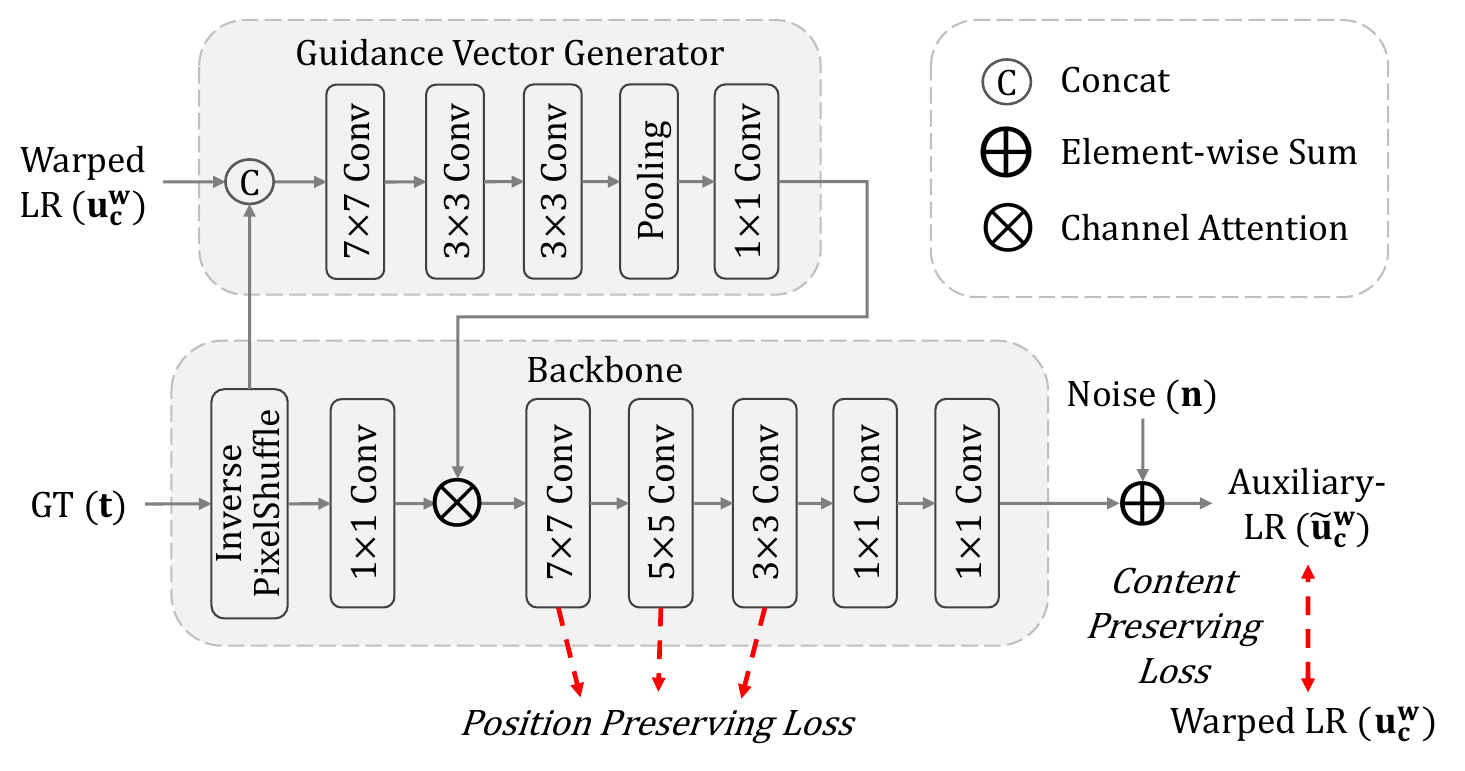}
    \end{overpic}
    \vspace{-2mm}
    \caption{Illustration of the auxiliary-LR generator. The position preserving loss constraints the kernel weight to ensure the alignment between auxiliary-LR and GT, while content preserving loss constraints that auxiliary-LR has similar contents and degradations as LR.}
    \label{fig:degrade}
    \vspace{-2mm}
  \end{figure}
  
  Thus, the auxiliary-LR $\mathbf{\tilde{u}_c^w}$ is required to satisfy two prerequisites.
  (i) $\mathbf{\tilde{u}_c^w}$ can be substituted by $\mathbf{u}$ during testing.
  (ii) The spatial position of $\mathbf{\tilde{u}_c^w}$ should keep the same as $\mathbf{t}$.
  For the first point, The auxiliary-LR should have similar contents and degradation types as LR, so that it can be substituted safely during testing.
  In particular, we design an auxiliary-LR generator network and constrain the contents of auxiliary-LR to be similar with these of LR, as shown in Fig.~\ref{fig:degrade}.
  For the second point, inspired by KernelGAN~\cite{KerGAN}, we take advantage of the position preserving loss to constrain the centroid of the convolution kernel in the center of space. 
  The position preserving loss $\mathcal{L}_\mathrm{p}$ can be defined as,
  \begin{equation}
    \begin{split}
    \mathcal{L}_\mathrm{p}(\mathbf{W^\mathit{l}}) &  =  \| \sum_{i=0}^{k-1}\sum_{j=0}^{k-1} (i- \frac{k}{2} + 0.5) \mathit{w_{i,j}^{l}} \|_1  \\
    &  + \| \sum_{i=0}^{k-1}\sum_{j=0}^{k-1} (j - \frac{k}{2} + 0.5)  \mathit{w_{i,j}^{l}} \|_1 ,
    \end{split}
    \label{eqn:loss_center}
  \end{equation}  
  where $\mathbf{W^\mathit{l}}$ denotes the kernel weight parameters of the $\mathit{l}$-th convolution layer in the backbone of the auxiliary-LR generator, $k$ is odd and denotes the kernel size, $\mathit{w_{i,j}^{l}}$ denotes the value in the $\mathit(i,j)$ position of $\mathbf{W^\mathit{l}}$.
  In addition, warped LR $\mathbf{u_c^w}$ can be used to generate a conditional guidance vector for modulating features of $\mathbf{t}$ globally, which does not affect the preservation of spatial position.
  Denote by $\mathcal{D}$ the auxiliary-LR generator, its optimization objective can be written as,
  \begin{equation}
    \Theta_\mathcal{D}^{*} = \arg \min_{\Theta_\mathcal{D}} \|\mathcal{D}(\mathbf{t}, \mathbf{u_c^w}; \Theta_\mathcal{D})  - \mathbf{u_c^w}\|_1  + \lambda_\mathit{p}\sum_{l=1}^L\mathcal{L}_\mathrm{p}(\mathbf{W^\mathit{l}}),
    \label{eqn:loss_d}
  \end{equation}
  where $\Theta_\mathcal{D}$ is generator's parameter and $\lambda_\mathit{p}$ is set to 100. 

  \changes{
  At this time, although the auxiliary-LR already has similar content and degradation as the LR in most cases (see Fig.~\ref{fig:misalignment}), it may struggle to cover some noise and artifacts that exist in LR image.
  Thus, the auxiliary-LR sometimes has clearer contents than the LR.
  When using these auxiliary-LR images to guide the alignment of the warped LR images, the restoration module may be overfitted to auxiliary-LR images, bringing adverse effects to the restoration of LR images. 
  To alleviate the problem, we add some simple perturbations (\eg, noise) to auxiliary-LR.
  Finally, the auxiliary-LR can be represented as,}
    \begin{equation}
      \mathbf{\tilde{u}_c} = \mathcal{D}(\mathbf{t}, \mathbf{u_c^w}; \Theta_\mathcal{D}^{*}) + \mathbf{n},
    \label{eqn:degrad}
    \end{equation}
  where $\mathbf{n}$ denotes Gaussian noise and JPEG compression noise.
  The variance of Gaussian noise is uniformly sampled from 5/255 to 30/255, and the JPEG quality factor is uniformly chosen from 60 to 95. 

\subsubsection{Aligning Warped LR to Auxiliary-LR} 
  \begin{figure}[t]
    \centering
    \begin{overpic} 
      [width=0.98\linewidth]{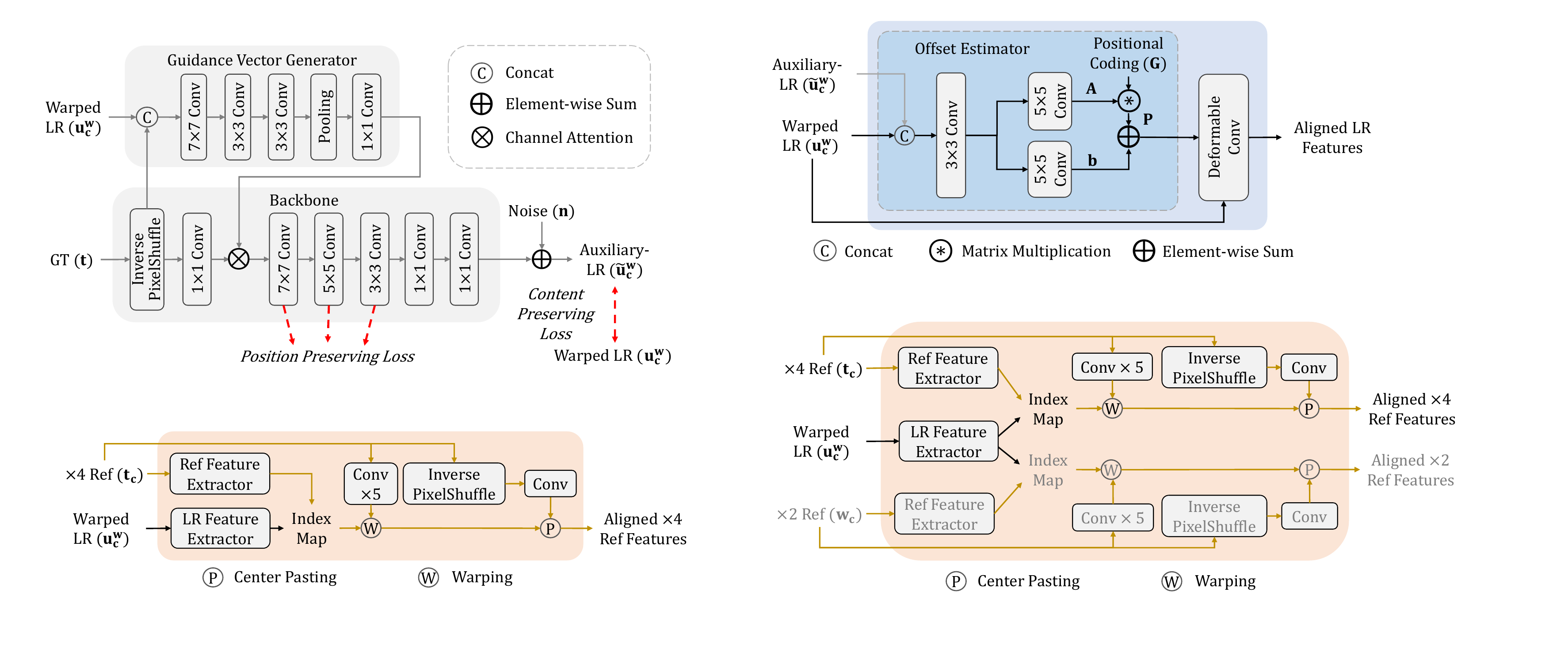}
    \end{overpic}
    \vspace{-1mm}
    \caption{Illustration of AdaSTN. The offset estimator predicts the position offsets between the warped LR and auxiliary-LR, then deformable convolution is used to deform warped LR according to the offsets.}
    \vspace{-3mm}
    \label{fig:AdaSTN}
  \end{figure}
  Given warped LR and auxiliary-LR, we suggest implicitly aligning warped LR to auxiliary-LR (aligned with GT).
  We can estimate the offsets between them and then deform the warped LR features to align with GT.
  Deformable convolution~\cite{DefConv} is a natural choice, but the direct estimation of the offsets may bring instability to the network training.
  Inspired by~\cite{AdaConv}, we utilize adaptive spatial transformer networks (AdaSTN) that offset is obtained indirectly by estimating the pixel-level affine transformation matrix and translation vector, as shown in Fig.~\ref{fig:AdaSTN}.
  For each pixel, the estimated offset of AdaSTN can be written as,
    \begin{equation}
      \mathbf{P} = \mathbf{A}\mathbf{G} + \mathbf{b}, 
    \label{eqn:offset}
    \end{equation}
  where $\mathbf{A}\in\mathbb{R}^{2\times2}$ is a predicted affine transformation matrix and $\mathbf{b}\in\mathbb{R}^{2\times1}$ is the translation vector.
  $\mathbf{G}$ is a positional coding represented by
    \begin{equation}
      \mathbf{G} = \begin{bmatrix}
                   -1 & -1 & -1 & 0  & 0 & 0 & 1  & 1 & 1 \\
                   -1 & 0  & 1  & -1 & 0 & 1 & -1 & 0 & 1
                   \end{bmatrix}.
    \label{eqn:G}
    \end{equation}
  Thus, the deformable convolution of AdaSTN can be formulated as,
    \begin{equation}
       \mathbf{y}(\mathbf{q}) = \sum\nolimits_{k=0}^8\mathbf{w}_k \mathbf{x}(\mathbf{q} + \mathbf{p}_k), 
    \label{eqn:defconv}
    \end{equation}
    where $\mathbf{x}$ and $\mathbf{y}$ represent the input and output features, respectively.
    $\mathbf{w}_k$ denotes kernel weight and $\mathbf{p}_k$ denotes  $k$-th column value of $\mathbf{P}$.
    %
    In experiments, we stack 3 AdaSTNs to align warped LR and auxiliary-LR progressively.
    
    Note that auxiliary-LR is not available in the testing phase.
    We can set $\mathbf{P} = \mathbf{0}$ directly, which means that the deformable convolution of AdaSTN can only observe the input value at the center point of the kernel and AdaSTN degenerates into 1$\times1$ convolution (see Fig.~\ref{fig:SelfDZSR}(c)).
    However, this way may produce some artifacts in the results due to the gap between training and testing.
    In order to bridge this gap, for each AdaSTN, we randomly set $\mathbf{P} = \mathbf{0}$ with probability $p$ (\eg, 0.3) during training.
    For each training sample, the probability $p^3$ (\eg, 0.027) that 3 AdaSTNs are all set to $\mathbf{P} = \mathbf{0}$ is low, so it has little impact on the learning of the overall framework.
    %
  
\subsection{Alignment between Ref and LR} \label{sec:alignment_ref}
  %
  %
  Previous RefSR methods generally perform matching by calculating cosine similarity between Ref and LR features.
  But for SelfDZSR during training, the misalignment between LR and GT will result in the warped Ref features being not aligned with GT after matching Ref to LR.
  Given that warped LR $\mathbf{\tilde{u}_c^w}$ is already roughly aligned with GT $\mathbf{t}$, we instead calculate the correlation between Ref and warped LR features (see Fig.~\ref{fig:SelfDZSR}(b)).
  During testing, the warped LR $\mathbf{\tilde{u}_c^w}$ can be substituted by the ultra-wide image $\mathbf{u}$ (see Fig.~\ref{fig:SelfDZSR}(c)).
  
\begin{figure}[t]
    \centering
    \begin{overpic} 
      [width=.98\linewidth]{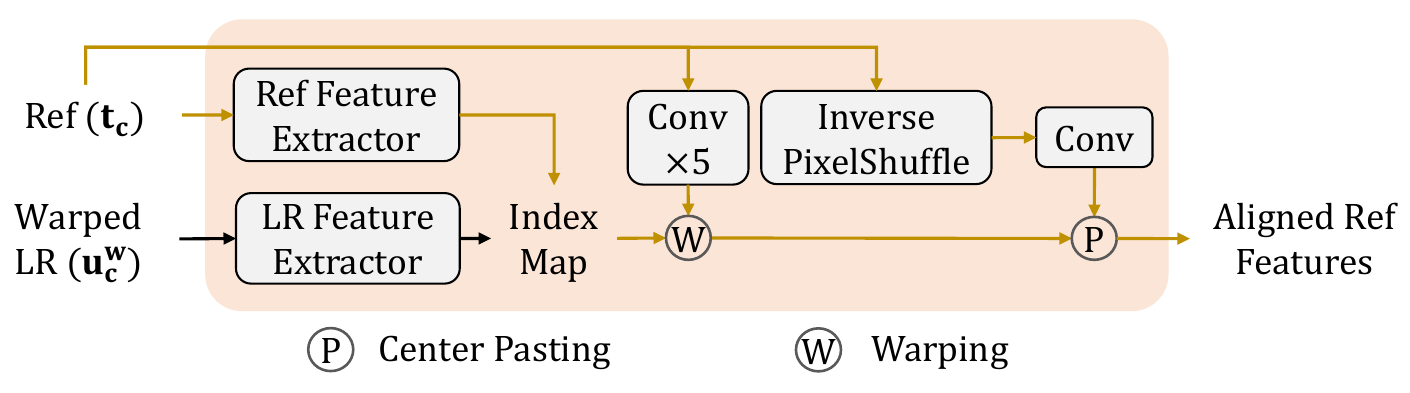}
    \end{overpic}
    \vspace{-2mm}
    \caption{Alignment between Ref and auxiliary-LR.}
    \vspace{-2mm}
    \label{fig:alignref}
  \end{figure}

  Fig.~\ref{fig:alignref} shows the detailed alignment scheme between Ref and warped LR.
  The index map is obtained by calculating the cosine similarity between Ref and warped LR features that are extracted by pre-trained feature extractors.
  Then the Ref is warped according to the index map.
  In addition, for SelfDZSR, the central part of LR has the same scene as Ref.
  Taking this property into account, we can rearrange Ref elements by an inverse PixelShuffle~\cite{PixelShuffle} layer, and then paste it to the center area of the warped Ref features.

  \begin{figure*}[t]
    \centering
    \begin{overpic} 
      [width=.95\linewidth]{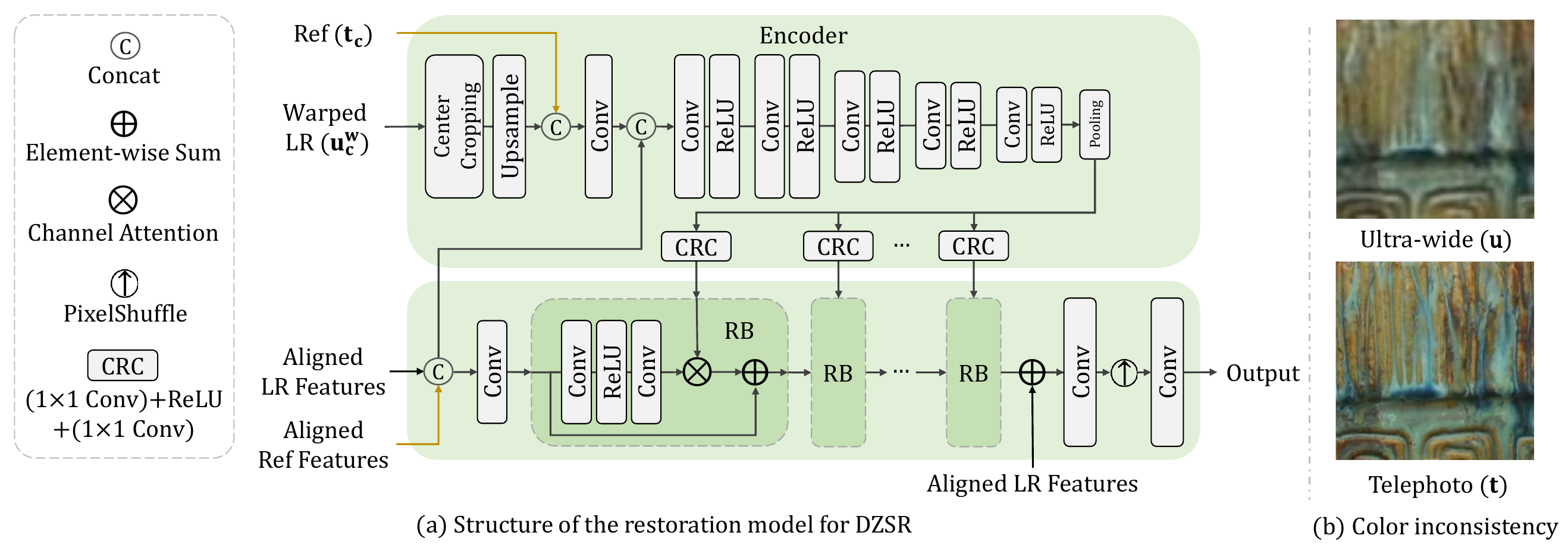}
    \end{overpic}
    \vspace{-2mm}
    \caption{(a) The detailed structure of the restoration model for DZSR. `RB' denotes the residual block~\cite{EDSR}.  (b) An example of color inconsistency between ultra-wide and telephoto images.}
    \vspace{-2mm}
    \label{fig:restor}
  \end{figure*}

\subsection{Restoration Module}  \label{sec:restoration}
  After getting the aligned LR features (introduced in Sec.~\ref{sec:alignment_lr}) and aligned Ref features (introduced in Sec.~\ref{sec:alignment_ref}), we feed them into the restoration module.
  Fig.~\ref{fig:restor}(a) shows the detailed structure of the restoration module.
  First, the aligned LR and aligned Ref features are concatenated and fed into the backbone, which consists of 16 residual blocks~\cite{EDSR}.
  Then the concatenated features, Ref image, and central area of warped LR image are input into an encoder to generate vectors that modulate the features of each residual block.
  This modulation can be regarded as channel attention on the features of the residual block.
  And it is beneficial to relieve the color inconsistency (see Fig.~\ref{fig:restor}(b)) between the real-world ultra-wide and telephoto images during testing.

\subsection{LOSW Loss and Learning Objective}~\label{sec:loss}
The sliced Wasserstein (SW) distance has exhibited  outstanding merit for training deep generative networks~\cite{SWDGen1,SWDGen2}.
  Recently, SW loss has been successfully applied in texture synthesis~\cite{SWLoss1}, image enhancement~\cite{SWLoss2}, image quality assessment~\cite{SWLossIQA} and \etc.
  And we also utilize SW loss to optimize our model in the earlier version SelfDZSR~\cite{SelfDZSR}.
  However, here we find that although it brings sharper results, it also leads to more artifacts.
  Thus, we improve SW loss and present local overlapped SW (LOSW) loss $\mathcal{L}_\mathrm{LOSW}$ to optimize SelfDZSR++ in this work.
  
  The algorithm of LOSW loss is described in Alg.~\ref{fig:pseudocode}.
  We first divide output and target VGG~\cite{VGGLOSS} features ($\mathbf{U}$ and $\mathbf{V}$) into overlapped small patches ($\mathbf{U_p}$ and $\mathbf{V_p}$), then obtain the patch representation  ($\mathbf{U_d}$ and $\mathbf{V_d}$) through random linear projection.
  Finally, we calculate the Wasserstein distance between the output and the target patch representation, which is defined as the element-wise $\ell_1$ distance over sorted patch representation ($\mathbf{U_s}$ and $\mathbf{V_s}$).
  LOSW loss and SW loss have different focuses in terms of feature similarity. Specifically, LOSW loss emphasizes the similarity of feature distribution in the local area, while SW loss focuses on that globally. Consequently, LOSW loss can encourage the output to be more faithful to the target image at the local level, and can also help reduce artifacts to some extent.
  SelfDZSR++ is jointly optimized with $\ell_1$ loss and LOSW loss. The total loss term can be written as,
  \begin{equation}
        \mathcal{L}_\mathrm{total}(\hat{\mathbf{y}}, \mathbf{t})
        = \|\hat{\mathbf{y}}-\mathbf{t}\|_1 
        + \lambda_\mathit{LOSW} \mathcal{L}_\mathrm{LOSW}(\phi(\hat{\mathbf{y}}), \phi(\mathbf{t})),
    \label{eqn:loss_selfDZSR}
  \end{equation}
  where $\phi$ denotes the pre-trained VGG-19~\cite{VGGLOSS} network, and we set $\lambda_\mathit{LOSW}=0.08$.

\begin{algorithm}[t]
  \caption{Pseudocode of LOSW loss}
  \label{fig:pseudocode}
  \hspace*{0.02in} {\bf Input:} $\mathbf{U}\in\mathbb{R}^{C \times K \times K}$: VGG features of output image;  \\
  \hspace*{0.43in} $\mathbf{V}\in\mathbb{R}^{C \times K \times K}$: VGG features of target image; \\
  \hspace*{0.43in} $\mathbf{M}\in\mathbb{R}^{C' \times C}$: random projection matrix; \\
  \hspace*{0.02in} {\bf Output:} $\mathcal{L}_\mathrm{SW}(\mathbf{U},\mathbf{V})$: the value of LOSW loss; 
  \begin{algorithmic}[1]
      \State Unfold features $\mathbf{U}$ and $\mathbf{V}$ to overlapped patch $\mathbf{U_p}(\in\mathbb{R}^{ P \times C \times k \times k})$ and $\mathbf{V_p}(\in\mathbb{R}^{P \times C \times k \times k})$ with kernel size $k$, respectively; $P$ denotes the number of patches;
      \State Flatten features $\mathbf{U_p}$ and $\mathbf{V_p}$ to $\mathbf{U_f}(\in\mathbb{R}^{P \times C \times k^2})$ and $\mathbf{V_f}(\in\mathbb{R}^{P \times C \times k^2})$, respectively;
      \State Project the features onto $C'$ directions: $\mathbf{U_d} = \mathbf{M}  \mathbf{U_f}$, $\mathbf{V_d} = \mathbf{M}  \mathbf{V_f}$;
      \State Sort projections for each direction: $\mathbf{U_s} = \mathbf{Sort}(\mathbf{U_d}$, dim=2), $\mathbf{V_s} = \mathbf{Sort}(\mathbf{V_d}$, dim=2);
      \State $\mathcal{L}_\mathrm{LOSW}(\mathbf{U},\mathbf{V}) = \|\mathbf{U_s}-\mathbf{V_s}\|_1 $
  \end{algorithmic}
\end{algorithm}

\subsection{Extension to Multiple Zoomed Observations}

  With the introduction in the previous sections (Sec.~\ref{sec:self-supervised} $\sim$ Sec.~\ref{sec:loss}), we can utilize dual zoomed observations (ultra-wide and telephoto images) for real-world RefSR in a self-supervised manner.
  In recent times, modern smartphones are being outfitted with not just two, but multiple lenses with different focal lengths. 
  This enables us to capture multiple images simultaneously with varying focal lengths. 
  Consequently, it is natural and significant to extend our method to multiple zoomed observations.
  In this subsection, we take image SR from triple zoomed observations (TZSR) as an example and introduce SelfTZSR++.
  Especially, we emphatically explore the fusion restoration scheme with Ref images at different focal lengths.

\noindent\textbf{Self-Supervised Learning for TZSR.}
  Some modern smartphones come equipped with three lenses that allow users to capture images at different focal lengths, including ultra-wide, wide-angle, and telephoto shots.
  As the focal length increases, the resolution of the image increases, while its field of view (FOV) gradually narrows.
  The proposed DZSR utilizes telephoto image $\mathbf{t}$ as a reference to super-resolve ultra-wide image $\mathbf{u}$.
  It's worth noting that the wide-angle image still has a higher resolution than the ultra-angle image, and potentially makes up for the lack of the telephoto image with a narrow FOV.
  To further improve SR results, TZSR aims to introduce the wide-angle image $\mathbf{w}$ as an additional reference, as shown in Fig~\ref{fig:intro}(c).
  For self-supervised training of TZSR, we introduce SelfTZSR++ based on SelfDZSR++.
  Specifically, SelfTZSR first crops the central area of the wide-angle image $\mathbf{w}$,
    \begin{equation}
      \mathbf{w_c} = \mathcal{C}(\mathbf{w}; r_w),
    \label{eqn:crop2}
    \end{equation}
  where $r_w$ is the focal length ratio between $\mathbf{w}$ and $\mathbf{u}$.
  Compared to the low-resolution image $\mathbf{u_c}$, the telephoto image $\mathbf{t_c}$ can be treated as a reference with a higher resolution of $\times r_t$, while the wide-angle image $\mathbf{w_c}$ can serve as another reference with a resolution of $\times r_w$.
  Then we can define TZSR as,
    \begin{equation}
      \Theta_\mathcal{Z} = \arg \min_{\Theta_\mathcal{Z}} \mathcal{L}\left(\mathcal{Z}(\mathbf{u_c}, \mathbf{w_c}, \mathbf{t_c}; \Theta_\mathcal{Z}), \mathbf{t} \right),
    \label{eqn:SelfTZSR}
    \end{equation}
  which is modified from Eqn.~(\ref{eqn:SelfDZSR}).
  SelfTZSR++ is also jointly optimized with $\ell_1$ loss and LOSW loss. The total loss term is the same as Eqn.~(\ref{eqn:loss_selfDZSR}).

  \begin{figure}[t]
    \centering
    \begin{overpic} 
      [width=.98\linewidth]{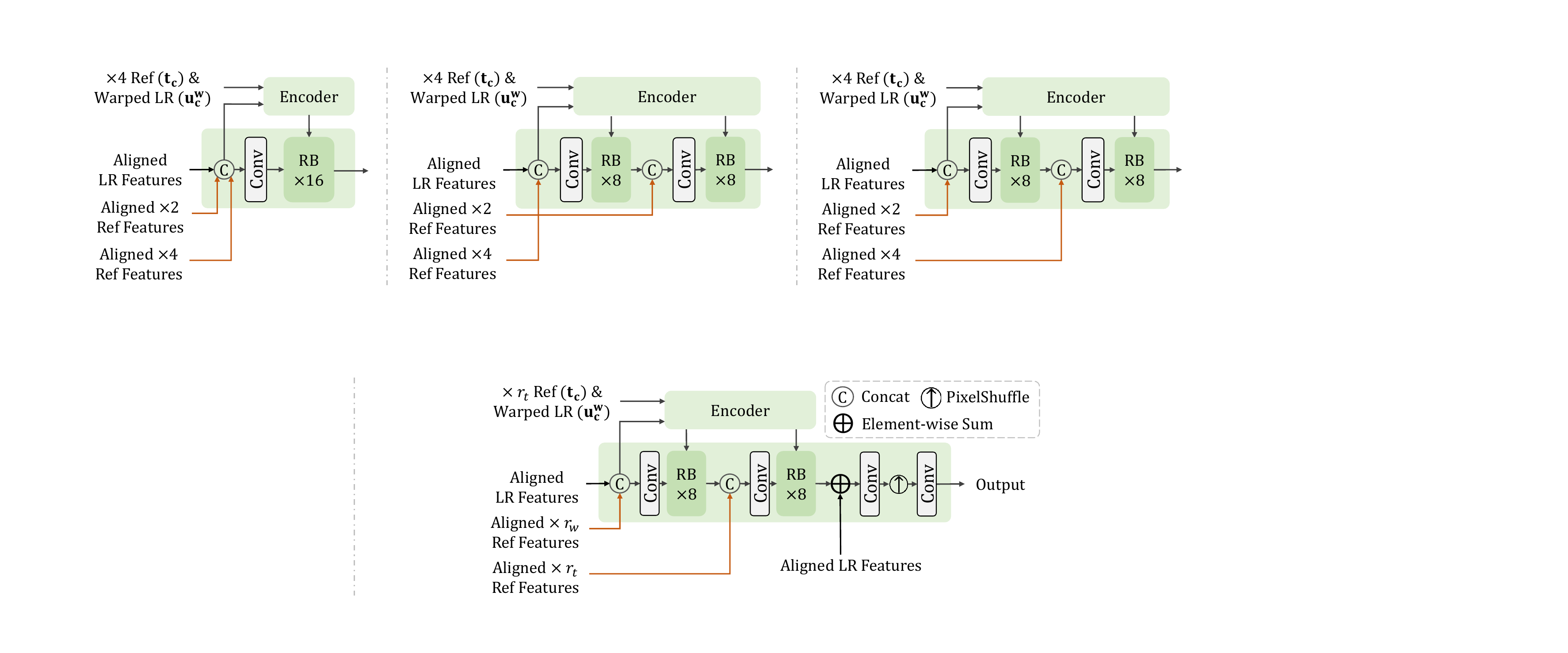}
    \end{overpic}
    \vspace{-2mm}
    \caption{Structure of the restoration model for TZSR. The aligned $\times r_w$ and $\times r_t$ features are from the central areas of the wide-angle $\mathbf{w}$ and telephoto $\mathbf{t}$ image, respectively. }
    \vspace{-2mm}
    \label{fig:tzsr}
  \end{figure}
  
\noindent\textbf{Progressive Fusion Restoration.}
  The aligned LR features, as well as the aligned $\times r_w$ Ref and $\times r_t$ Ref features can be obtained following the methods introduced in Sec.~\ref{sec:alignment_lr} and Sec.~\ref{sec:alignment_ref}, respectively.
  Here, we focus on how these features can be processed for better restoration.
  Perhaps due to the information bottleneck, it can not achieve the best performance when directly concatenating these features to feed the restoration network.
  Instead, we propose a progressive fusion scheme in which aligned LR features are sequentially fused with two Ref features.
  Specifically, as shown in Fig.~\ref{fig:tzsr}, we first concatenate $\times r_w$ Ref features (which have a lower resolution than the $\times r_t$ Ref ones) with the LR ones, and process them. 
  Then we merge $\times r_t$ Ref features with the processed features for further modulation.
  Through the progressive utilization of Refs with different resolutions, the effect of image SR can be gradually improved.

\begin{table*}[ht] 
  \small
  \caption{Quantitative results of SR models trained only with $\ell_1$ (or $\ell_2$) loss. The best and second-best results are masked by {\color{red}red} and {\color{blue}blue} colors, respectively. RefSR$^\dagger$ represents that the RefSR methods are trained in our self-supervised learning manner.}
  \label{tab:all-l1}
  \centering\noindent
  \centering%
  \vspace{-7mm}
  \begin{center}
    \begin{tabular}{clcccc}
      \toprule
      & \multirow{3}{*}{Method}
      & \multicolumn{4}{c}{{PSNR}{$\uparrow$} {/} {SSIM}{$\uparrow$} {/} {LPIPS}{$\downarrow$}} \\
      & & \multicolumn{2}{c}{\textbf {Nikon Camera Images}}
      & \multicolumn{2}{c}{\textbf {Iphone Camera Images}} \\
      &  & \emph{Full-Image}  & \emph{Corner-Image} & \emph{Full-Image}  & \emph{Corner-Image} \\
        \midrule
        \multirow{3}{*}{SISR}
        & EDSR~\cite{EDSR}   & 27.26 / 0.8364 / 0.362 & 27.29 / 0.8345 / 0.363 & 23.54 / 0.7224 / 0.376  &  23.54 / 0.7236 / 0.361  \\ 
        & RCAN~\cite{RCAN}   & 27.30 / 0.8344 / 0.383 & 27.33 / 0.8323 / 0.383 & 23.51 / 0.7249 / 0.376  &  23.50 / 0.7261 / 0.361   \\ 
        & CDC~\cite{CDC}    & 27.20 / 0.8306 / 0.412 & 27.24 / 0.8283 / 0.412 & 23.49 / 0.7167 / 0.433  &  23.49 / 0.7177 / 0.417   \\ 
        \midrule
        \multirow{6}{*}{RefSR$^\dagger$}  
        & SRNTT-$\ell_2$~\cite{SRNTT}    & 27.30 / 0.8387 / 0.359 & 27.33 / 0.8366 / 0.359 & 23.60 / 0.7194 / 0.419  &  23.59 / 0.7205 / 0.403  \\ 
        & TTSR-$\ell_1$~\cite{TTSR}    & 25.83 / 0.8272 / 0.369 & 25.80 / 0.8259 / 0.369 & 23.48 / 0.7199 / 0.370 &  23.47 / 0.7210 / 0.355 \\ 
        & $C^2$-Matching-$\ell_1$~\cite{C2-Matching}  & 27.19 / 0.8402 / 0.362 & 27.23 / 0.8381 / 0.362 & 23.51 / 0.7175 / 0.391  &  23.51 / 0.7185 / 0.375 \\ 
        & MASA-$\ell_1$~\cite{MASA-SR}   & 27.27 / 0.8372 / 0.339 & 27.30 / 0.8352 / 0.339 & 23.53 / 0.7196 / 0.391  &  23.51 / 0.7208 / 0.375  \\ 
        & DCSR-$\ell_1$~\cite{DCSR}   & 27.73 / 0.8274 / 0.355 & 27.72 / 0.8275 / 0.349 &   23.23 / 0.7173 / 0.383  &  23.22 / 0.7186 / 0.367  \\
        \midrule
        \multirow{3}{*}{Ours}  
        & SelfDZSR-$\ell_1$~\cite{SelfDZSR}  &  28.93 / 0.8572 / 0.308 & 28.67 / 0.8457 / 0.328 & 23.79 / 0.7396 / {\color{blue}0.320}  &  23.52 / 0.7243 / {\color{blue}0.325}  \\ 
        & SelfDZSR++-$\ell_1$   & {\color{blue}29.63} / {\color{blue}0.8663} / {\color{blue}0.290} & {\color{blue}29.36} / {\color{blue}0.8555} / {\color{blue}0.309} & {\color{blue}24.08} / {\color{blue}0.7502} / 0.332  &  {\color{blue}23.81} / {\color{blue}0.7347} / 0.338 \\ 
        & SelfTZSR++-$\ell_1$   & {\color{red}29.74} / {\color{red}0.8708} / {\color{red}0.280} & {\color{red}29.47} / {\color{red}0.8617} / {\color{red}0.295} &  {\color{red}24.36} / {\color{red}0.7680} / {\color{red}0.303} & {\color{red}24.10} / {\color{red}0.7576} / {\color{red}0.305} \\
      \bottomrule
    \end{tabular}
    \end{center}
\end{table*}  

\begin{table*}[ht] 
  \small
  \caption{Quantitative results of SR models trained with their all loss terms. The best and second-best results are masked by {\color{red}red} and {\color{blue}blue} colors, respectively. RefSR$^\dagger$ represents that the RefSR methods are trained in our self-supervised learning manner.}
  \label{tab:all-loss}
  \centering\noindent
  \centering%
  \vspace{-7mm}
  \begin{center}
    \begin{tabular}{clcccc}
      \toprule
      & \multirow{3}{*}{Method}
      & \multicolumn{4}{c}{{PSNR}{$\uparrow$} {/} {SSIM}{$\uparrow$} {/} {LPIPS}{$\downarrow$}} \\
      & & \multicolumn{2}{c}{\textbf {Nikon Camera Images}}
      & \multicolumn{2}{c}{\textbf {Iphone Camera Images}} \\
      &  & \emph{Full-Image}  & \emph{Corner-Image} & \emph{Full-Image}  & \emph{Corner-Image} \\
        \midrule
        \multirow{2}{*}{SISR}
        & BSRGAN~\cite{BSRGAN} & 26.91 / 0.8151 / 0.279  &  26.96 / 0.8135 / 0.278 & 22.15 / 0.6833 / 0.313  &  22.14 / 0.6844 / 0.298   \\ 
        & Real-ESRGAN~\cite{Real-ESRGAN}  & 25.96 / 0.8076 / 0.272 & 26.00 / 0.8063 / 0.271 & 21.78 / 0.6847 / 0.311  &  21.77 / 0.6859 / 0.296   \\ 
        \midrule
        \multirow{6}{*}{RefSR$^\dagger$}  
        & SRNTT~\cite{SRNTT}           & 27.31 / 0.8242 / 0.286 & 27.35 / 0.8223 / 0.283  & 23.29 / 0.6833 / 0.349  &  23.28 / 0.6847 / 0.334  \\ 
        & TTSR~\cite{TTSR}         & 25.31 / 0.7719 / 0.282 & 25.27 / 0.7708 / 0.282 & 22.40 / 0.6514 / 0.342  &  22.39 / 0.6528 / 0.325   \\ 
        & $C^2$-Matching~\cite{C2-Matching}  & 26.79 / 0.8141 / 0.327 & 26.81 / 0.8123 / 0.325 & 22.57 / 0.6725 / 0.349  &  22.56 / 0.6734 / 0.331  \\
        & MASA~\cite{MASA-SR}           & 27.32 / 0.7640 / 0.273  & 27.37 / 0.7615 / 0.274 & 21.75 / 0.6277 / 0.324  &  21.75 / 0.6291 / 0.308  \\
        & DCSR~\cite{DCSR}              & 27.69 / 0.8232 / 0.276 & 27.68 / 0.8232 / 0.272 &  23.08 / 0.7014 / 0.308 & 23.07 / 0.7029 / 0.294  \\
        \midrule
        \multirow{3}{*}{Ours}  
        & SelfDZSR~\cite{SelfDZSR}  & 28.67 / 0.8356 / 0.219 & 28.42 / 0.8238 / 0.231 & 23.37 / 0.7128 / 0.252  &  23.12 / 0.6987 / 0.250  \\ 
        & SelfDZSR++     & {\color{blue}29.30} / {\color{blue}0.8511} / {\color{blue}0.201} & {\color{blue}29.03} / {\color{blue}0.8401} / {\color{blue}0.213} & {\color{blue}23.64} / {\color{blue}0.7266} / {\color{blue}0.244}  &  {\color{blue}23.38} / {\color{blue}0.7105} / {\color{blue}0.247}  \\ 
        & SelfTZSR++    & {\color{red}29.62} / {\color{red}0.8582} / {\color{red}0.187} & {\color{red}29.37} / {\color{red}0.8490} / {\color{red}0.196} & {\color{red}24.00} / {\color{red}0.7466} / {\color{red}0.215} & {\color{red}23.76} / {\color{red}0.7358} / {\color{red}0.213}  \\  
      \bottomrule
    \end{tabular}
    \end{center}
    \vspace{-3mm}
\end{table*}

\begin{figure*}[ht]
    \begin{minipage}[t]{\linewidth}
        \centering
        \subfloat[Ultra-wide]
        {
            \includegraphics[height=.142\linewidth, width=.213\linewidth]{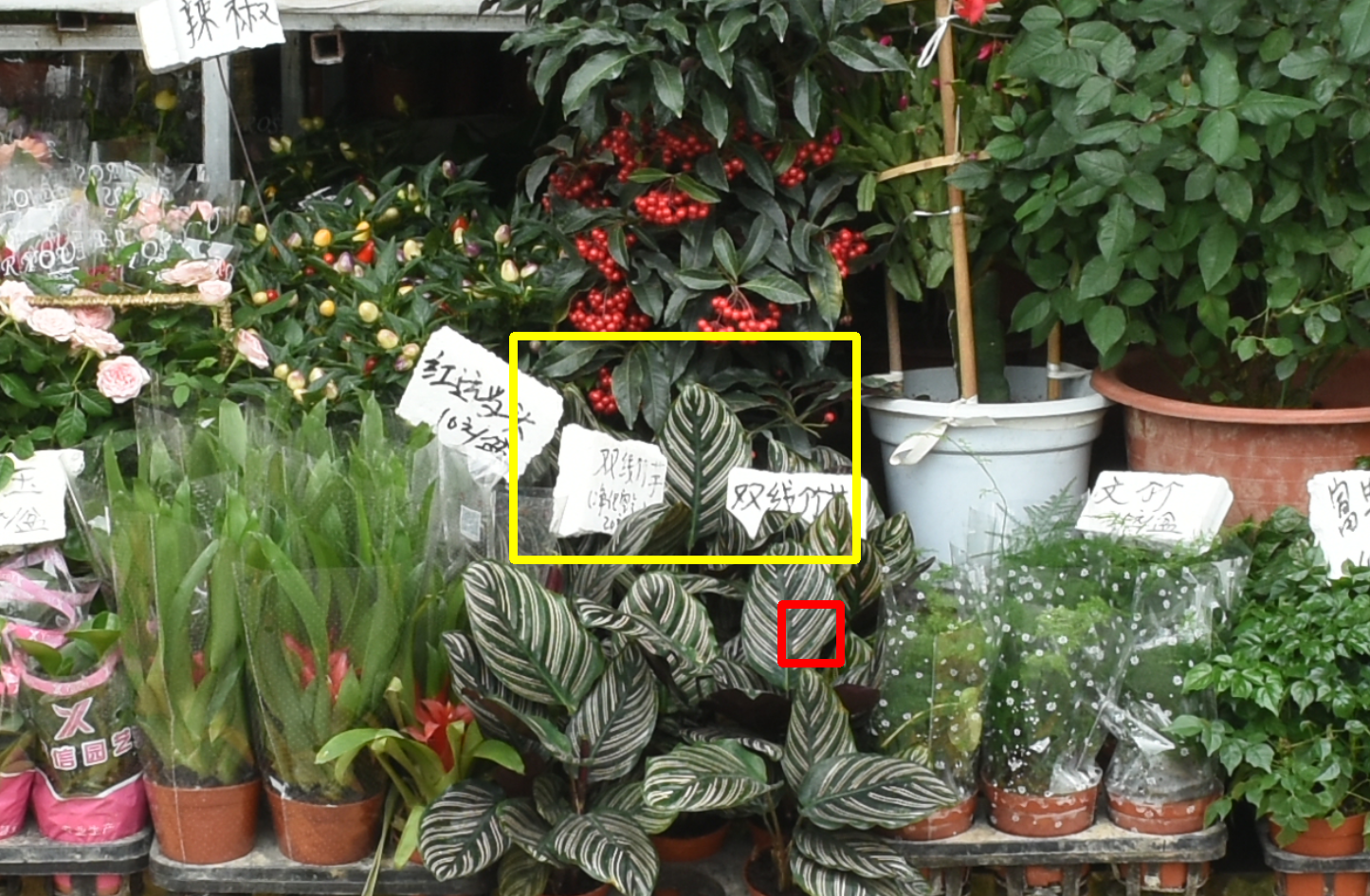}
        }
        \subfloat[LR]
        {
            \includegraphics[height=.142\linewidth, width=.142\linewidth]{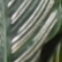}
        }
        \subfloat[Real-ESRGAN~\cite{Real-ESRGAN}]
        {
            \includegraphics[height=.142\linewidth,width=.142\linewidth]{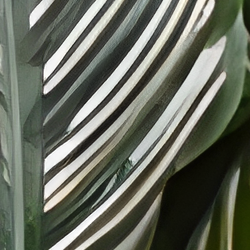}
        }
        \subfloat[TTSR\cite{TTSR}]
        {
            \includegraphics[height=.142\linewidth,width=.142\linewidth]{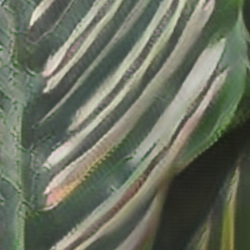}
        }
        \subfloat[$C^2$-Matching\cite{C2-Matching}]
        {
            \includegraphics[height=.142\linewidth,width=.142\linewidth]{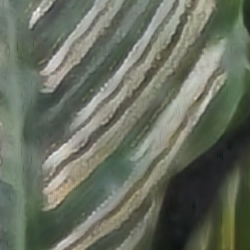}
        }
        \subfloat[MASA\cite{MASA-SR}]
        {
            \includegraphics[height=.142\linewidth,width=.142\linewidth]{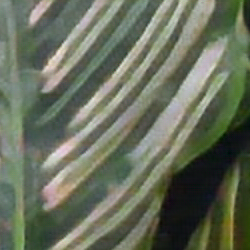}
        }

        \vspace{-2mm}
        \subfloat[][Telephoto]
        {
            \includegraphics[height=.142\linewidth,height=.142\linewidth, width=.213\linewidth]{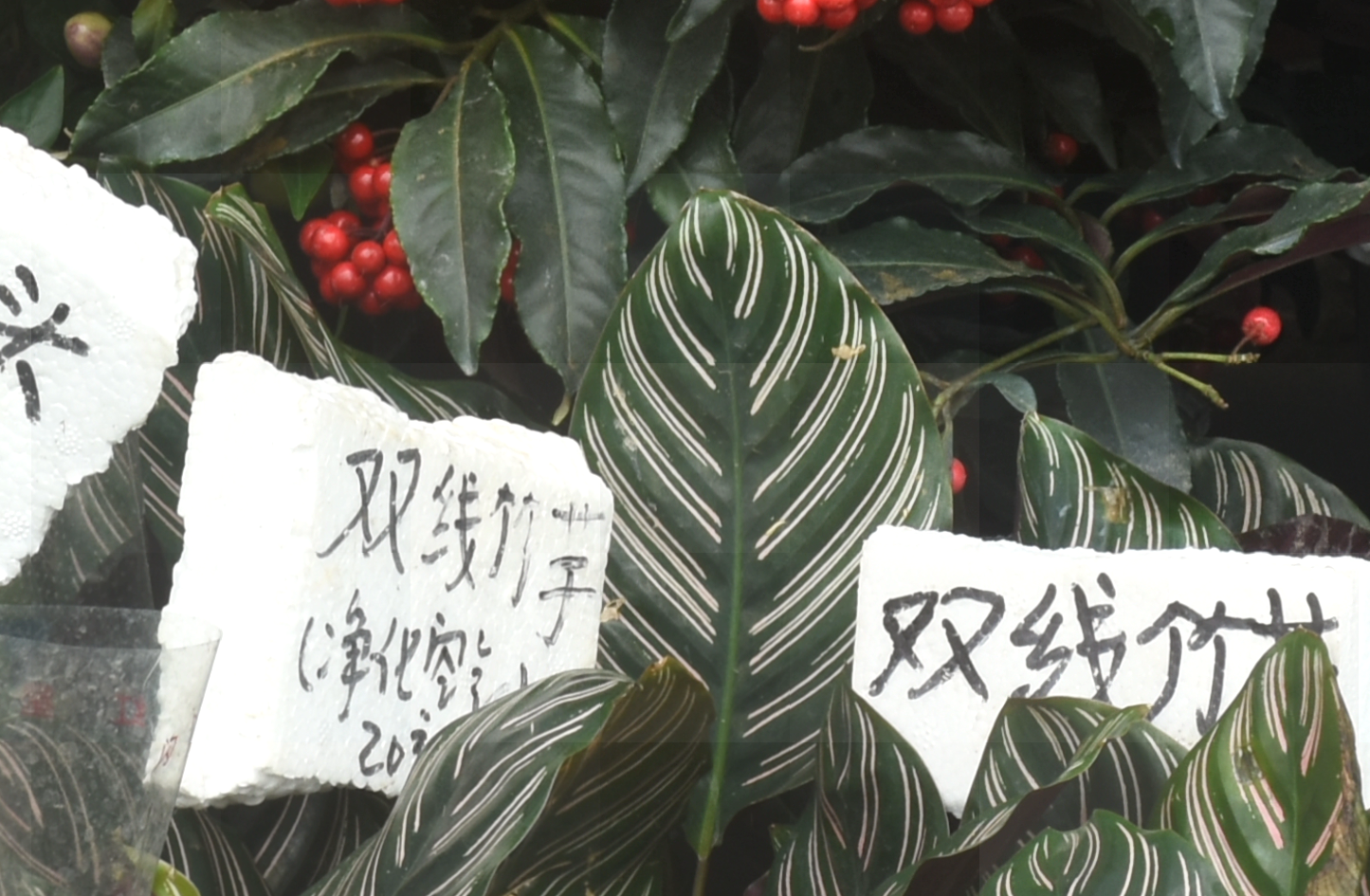}
        }
        \subfloat[DCSR\cite{DCSR}]
        {
            \includegraphics[height=.142\linewidth,width=.142\linewidth]{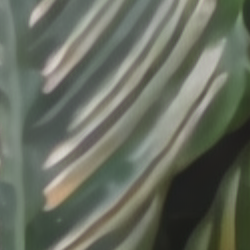}
        }
        \subfloat[SelfDZSR~\cite{SelfDZSR}]
        {
            \includegraphics[height=.142\linewidth,width=.142\linewidth]{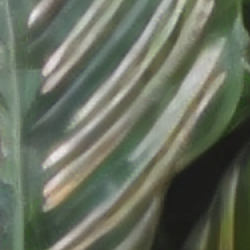}
        }
        \subfloat[SelfDZSR++]
        {
            \includegraphics[height=.142\linewidth,width=.142\linewidth]{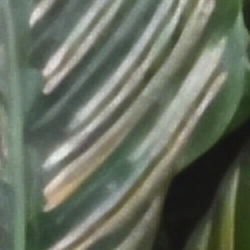}
        }
        \subfloat[SelfTZSR++]
        {
            \includegraphics[height=.142\linewidth,width=.142\linewidth]{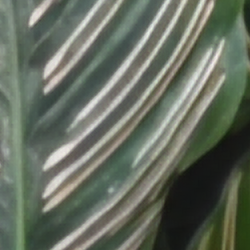}
        }
        \subfloat[GT]
        {
            \includegraphics[height=.142\linewidth,width=.142\linewidth]{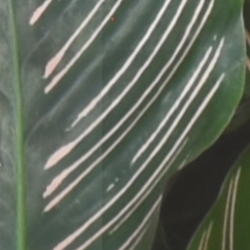}
        }
    \end{minipage}
    \vspace{-2mm}
    \caption{Visual comparison on Nikon camera images. In the ultra-wide image, the yellow box indicates the overlapped scene with the telephoto image, while the red box represents the selected LR patch.}
    \label{fig:Nikon}
    \vspace{-4mm}
\end{figure*}

\begin{figure*}[ht]
    \begin{minipage}[t]{\linewidth}
        \centering
        \subfloat[Ultra-wide]
        {
            \includegraphics[height=.142\linewidth, width=.213\linewidth]{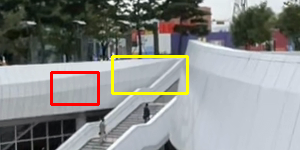}
        }
        \subfloat[LR]
        {
            \includegraphics[height=.142\linewidth, width=.142\linewidth]{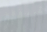}
        }
        \subfloat[Real-ESRGAN~\cite{Real-ESRGAN}]
        {
            \includegraphics[height=.142\linewidth,width=.142\linewidth]{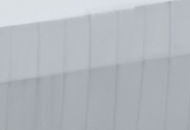}
        }
        \subfloat[TTSR\cite{TTSR}]
        {
            \includegraphics[height=.142\linewidth,width=.142\linewidth]{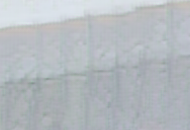}
        }
        \subfloat[$C^2$-Matching\cite{C2-Matching}]
        {
            \includegraphics[height=.142\linewidth,width=.142\linewidth]{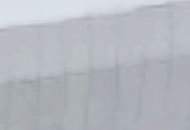}
        }
        \subfloat[MASA\cite{MASA-SR}]
        {
            \includegraphics[height=.142\linewidth,width=.142\linewidth]{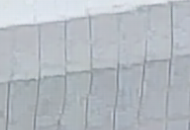}
        }

        \vspace{-2mm}
        \subfloat[][Telephoto]
        {
            \includegraphics[height=.142\linewidth,height=.142\linewidth, width=.213\linewidth]{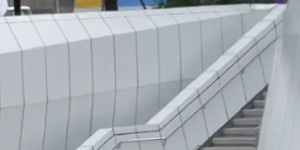}
        }
        \subfloat[DCSR\cite{DCSR}]
        {
            \includegraphics[height=.142\linewidth,width=.142\linewidth]{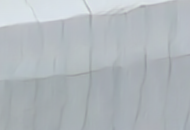}
        }
        \subfloat[SelfDZSR~\cite{SelfDZSR}]
        {
            \includegraphics[height=.142\linewidth,width=.142\linewidth]{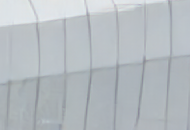}
        }
        \subfloat[SelfDZSR++]
        {
            \includegraphics[height=.142\linewidth,width=.142\linewidth]{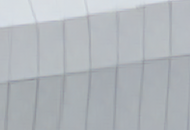}
        }
        \subfloat[SelfTZSR++]
        {
            \includegraphics[height=.142\linewidth,width=.142\linewidth]{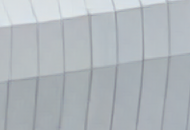}
        }
        \subfloat[GT]
        {
            \includegraphics[height=.142\linewidth,width=.142\linewidth]{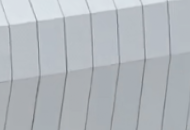}
        }
    \end{minipage}
    \vspace{-2mm}
    \caption{Visual comparison on iPhone camera images. In the ultra-wide image, the yellow box indicates the overlapped scene with the telephoto image, while the red box represents the selected LR patch.}
    \label{fig:iPhone} 
    \vspace{-2mm}
\end{figure*}

\begin{table}[t!] 
  \vspace{1mm}
  \small
  \centering
  \caption{Model \#parameters and \#FLOPs comparison of SISR and RefSR methods. The \#FLOPs is measured when $\times4$ super-resolving LR image to  $1280\times720$ resolution. For RefSR methods, the Ref image has the same size with LR.} %
  \label{tab:flops}
  \centering\noindent
  \centering%
  \vspace{-3mm}
  \begin{center}
    \begin{tabular}{clcc}
      \toprule
      & Method &  \# Params (M) & \#FLOPs (G) \\
      \midrule
      \multirow{5}{*}{SISR}
      & EDSR~\cite{EDSR}  & 43.1 & 5792 \\
      & RCAN~\cite{RCAN}  & 15.6 & 1838   \\
      & CDC~~\cite{CDC} & 39.9  & 1626  \\
      & BSRGAN~\cite{BSRGAN} & 16.7  & 2068  \\
      & Real-ESRGAN~\cite{Real-ESRGAN} &  16.7 & 2068  \\
      \midrule
      \multirow{5}{*}{RefSR}  
      & SRNTT~\cite{SRNTT}  & 5.5 & 3568\\ 
      & TTSR~\cite{TTSR}  & 7.3 & 2468 \\ 
      & $C^2$-Matching~\cite{C2-Matching}  & 8.9 & 1968 \\ 
      & MASA~\cite{MASA-SR}  & 4.0 & 1984  \\ 
      & DCSR~\cite{DCSR} & 3.2 & 836 \\ 
      \midrule
      \multirow{3}{*}{Ours}
      & SelfDZSR~\cite{SelfDZSR}   & 3.1 & 454 \\ 
      & SelfDZSR++  & 3.1 & 454   \\ 
      & SelfTZSR++  & 3.3 & 538  \\ 
      \bottomrule  
    \end{tabular}
    \vspace{-5mm}
    \end{center}
\end{table}

\section{Experiments} \label{sec:experiments}

\subsection{Experimental Setup}
  {\textbf{Datasets.}}
  Experiments are conducted on Nikon camera images from DRealSR dataset~\cite{CDC} and iPhone camera images from RefVSR dataset~\cite{RefVSR}.
  The training patches of DRealSR have been manually and carefully selected for mitigating the alignment issue, which is laborious and time-consuming.
  Instead, we take the originally captured images for training, making the whole process fully automated.
  In particular, each scene of the original data contains four different focal-length images. 
  We adopt the shortest focal-length image, the second shortest focal-length one, and the longest focal-length one as the ultra-wide, wide-angle, and telephoto images, respectively.
  There are 163 image pairs for training and 20 for evaluation.
  RefVSR~\cite{DCSR} dataset is collected by iPhone 12 Pro Max, which provides three videos with different fixed focal lengths (ultra-wide, wide-angle, and telephoto) for each scene.
  We remove blurry frames, and treat the video frames as single images.
  There are 13893 image pairs for training and 1024 for evaluation.
  For simplicity, we resize the above images so that $r_w$ and $r_t$ are 2 and 4, respectively.

  \noindent\textbf{Training Configurations.}
  We augment the training data with random horizontal flip, vertical flip and $90^\circ$ rotation.
  The batch size is 16, and the patch size for LR is 48$\times$48.
  The model is trained with the Adam optimizer~\cite{Adam} by setting $\beta_1=0.9$ and $\beta_2=0.999$ for 400 epochs. 
  The learning rate is initially set to $1\times10^{-4}$ and is decayed to $5\times10^{-5}$ after 200 epochs.
  The experiments are conducted with PyTorch~\cite{PyTorch} on an Nvidia GeForce RTX 3090 GPU.

 \noindent\textbf{Evaluation Configurations.}
  No ground-truths can be utilized when inputting the ultra-wide and telephoto images directly, leading to quantitative evaluation difficult.
  As a result, we still use the center areas of ultra-wide and telephoto images as LR and Ref, respectively. Then we align the whole telephoto image with the output by optical flow network~\cite{PWC-Net}.
  Quantitative metrics (\ie, PSNR, SSIM~\cite{SSIM} and LPIPS~\cite{LPIPS}) can be calculated between the output and the aligned telephoto image.
  Noted that the scene of the Ref is the center area of the telephoto image.
  In addition to calculating the metrics on the full image (marked as \emph{Full-Image}), we also calculate the metrics of the areas excluding the center (marked as \emph{Corner-Image}).
  And all patches for visual comparison are selected from the areas excluding the center of the output. 

\subsection{Quantitative and Qualitative Results}
  We compare results with SISR (\ie, EDSR~\cite{EDSR}, RCAN~\cite{RCAN}, CDC~\cite{CDC}, BSRGAN~\cite{BSRGAN} and Real-ESRGAN~\cite{Real-ESRGAN}) and RefSR (\ie, SRNTT~\cite{SRNTT}, TTSR~\cite{TTSR}, MASA~\cite{MASA-SR}, $C^2$-Matching~\cite{C2-Matching}, DCSR~\cite{DCSR} and our SelfDZSR~\cite{SelfDZSR}) methods.
  The results of BSRGAN and Real-ESRGAN are generated via the officially released model, other methods are retrained using our images for a fair comparison.
  Among them, RefSR methods are trained in our self-supervised learning manner and each method has two models, obtained by minimizing $\ell_1$ (or $\ell_2$) loss and all loss terms that are used in their papers.

  Table~\ref{tab:all-l1} and Table~\ref{tab:all-loss} show the quantitative results of SR models trained with $\ell_1$ (or $\ell_2$) loss and their all loss terms, respectively.  
  From the tables, SelfDZSR~\cite{SelfDZSR} has exceeded most previous SISR and RefSR methods, benefiting from the alignment of data pairs and effective utilization of Ref information.
  Due to the better handling of the alignment problem and the proposal of LOSW loss, SelfDZSR++ outperforms SelfDZSR on most metrics.
  Moreover, with the further introduction of additional Ref and progressive fusion scheme, SelfTZSR++ exceeds all competing methods both in terms of fidelity and perception.

  The visual comparison on Nikon and iPhone camera images can be seen in Fig.~\ref{fig:Nikon} and Fig.~\ref{fig:iPhone}, respectively.
  Our results usually restore more fine-scale textures, and are clearer and more photo-realistic.

\subsection{Comparison of \#Parameters and \#FLOPs} 
We also compare the number of parameters and FLOPs of different models, as shown in Table~\ref{tab:flops}.
For RefSR methods, the cost of calculating the similarity between LR and Ref occupies a large part of the computational cost.
In this work, we calculate cosine similarity between $\times4$ down-sampled Ref and $\times4$ down-sampled LR features, and find that its performance is close to that of computing similarity at the original image size.
By virtue of the lightweight restoration model and the fast similarity calculation, our method has low \#parameters and \#FLOPs in comparison to both SISR and RefSR methods.

\section{Ablation Study}
\label{sec:ablation}
  In this section, we conduct ablation experiments for assessing the effect of self-supervised learning, alignment between LR and GT, LOSW loss, different Refs, and fusion scheme.
  Unless otherwise stated, experiments are carried out on the Nikon camera images~\cite{CDC} with SelfTZSR++, and the metrics are evaluated on full images.
\subsection{Effect of Self-Supervised Learning}
\begin{figure}[t]
	\centering
	\subfloat[Bicubic]
	{
		\begin{minipage}{0.23\linewidth}
			\centering
			\includegraphics[width=\linewidth]{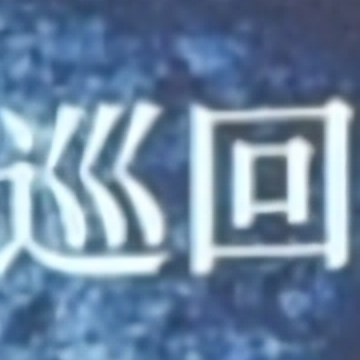}
		\end{minipage}
	}
	\hspace{-0.1cm}
	\subfloat[SRA\cite{DCSR}] 
	{
		\begin{minipage}{0.23\linewidth}
			\centering
			\includegraphics[width=1\linewidth]{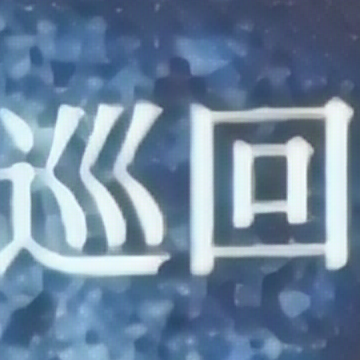}
		\end{minipage}
	}
	\hspace{-0.1cm}
	\subfloat[Ours] 
	{
		\begin{minipage}{0.23\linewidth}
			\centering
			\includegraphics[width=\linewidth]{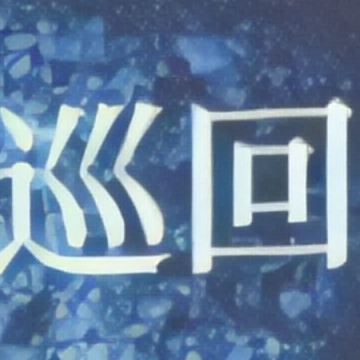}
		\end{minipage}
	}
	\hspace{-0.1cm}
	\subfloat[GT]
	{
		\begin{minipage}{0.23\linewidth}
			\centering
			\includegraphics[width=\linewidth]{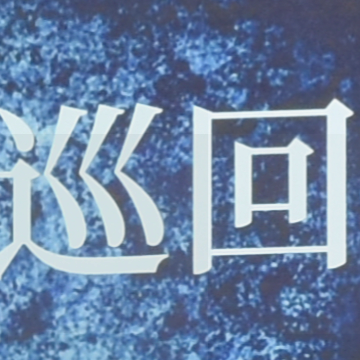}
		\end{minipage}
	}%
  \vspace{-2mm}
  \caption{Visual result comparison when using different training strategies. Our result is sharper and clearer.}
  \label{fig:ablation_self}
  \vspace{-2mm}
\end{figure}

  In order to verify the effectiveness of our proposed self-supervised approach (see Sec.~\ref{sec:self-supervised}), we conduct experiments on different training strategies.
  First, we remove the two-stage alignment components and AdaSTN in SelfDZSR++.
  Then we replace the real-world LR image with the bicubic downsampling GT image, and retrain the network.
  Finally, for a fair comparison, we take the self-supervised real-image adaptation (SRA)~\cite{DCSR} strategy and our self-supervised method to fine-tune the above model, respectively.
  As can be seen from Table~\ref{tab:ablation_self}, when evaluating on real-world images, our proposed self-supervised method achieves better results. 
  The PSNR metric is 1.03 dB higher than the model based on SRA fine-tuning.
  From Fig.~\ref{fig:ablation_self}, our visual result is sharper and clearer. 
  In a word, it can be seen that even if the misalignment between LR and GT is not handled, our self-supervised method is still better than SRA~\cite{DCSR} strategy.

\begin{figure}[t]
    \centering
    \begin{overpic}
      [width=\linewidth]{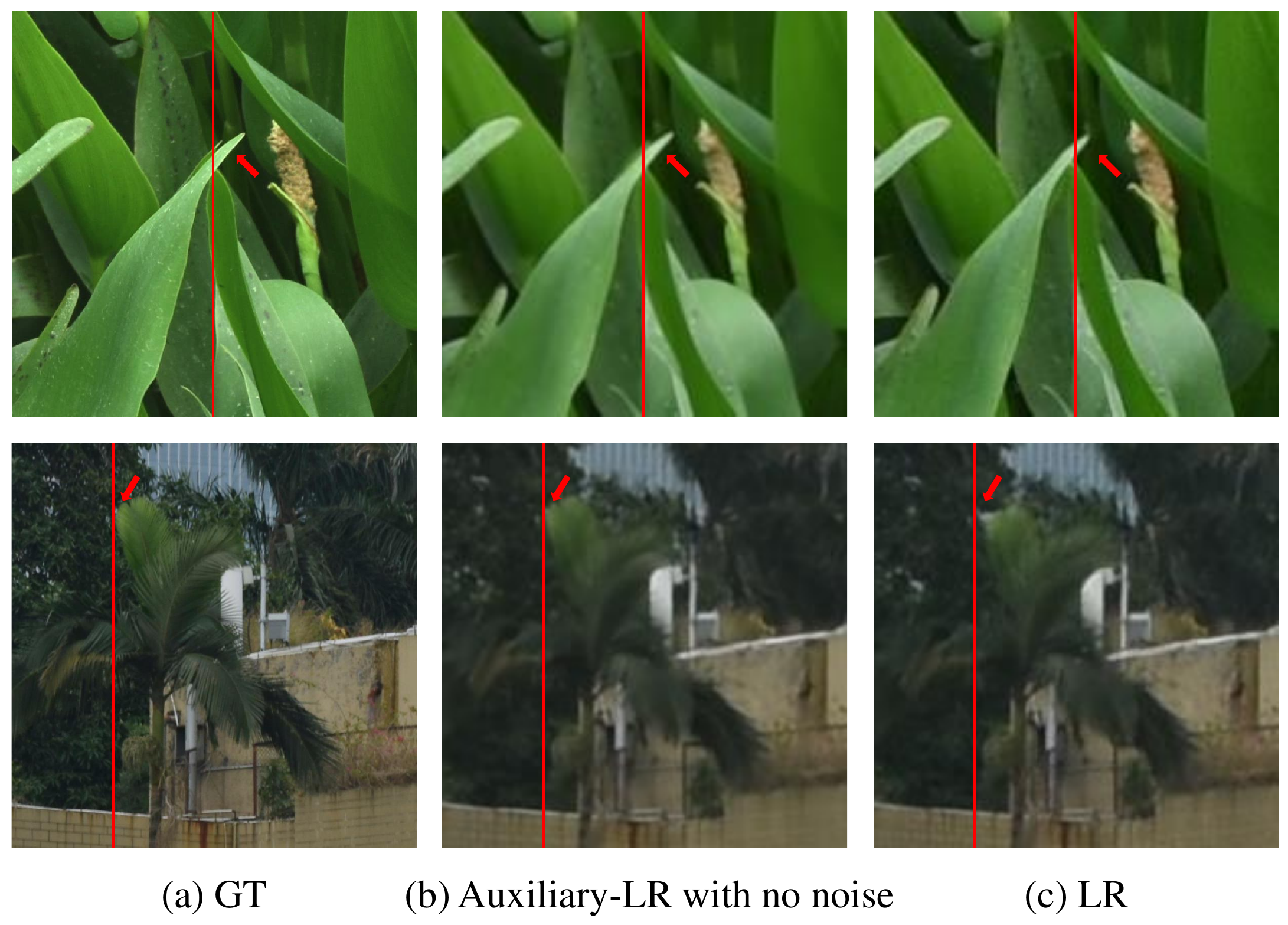}
    \end{overpic}
    \vspace{-6mm}
    \caption{Visual results of noisy-free auxiliary-LR image. The auxiliary-LR has similar contents as LR and is aligned with GT. The red lines and arrows in the same row are in the same position relative to the image.}
    \label{fig:misalignment}
    \vspace{-2mm}
\end{figure}

\begin{table}[t] 
  \small
  \setlength{\tabcolsep}{1mm}
  \caption{Ablation study on training strategies.}
  \label{tab:ablation_self}
  \centering\noindent
  \centering%
  \vspace{-4mm}
  \begin{center}
    \begin{tabular}{lc}
      \toprule  
      {Training Strategy\quad} & 
      {PSNR}{$\uparrow$} / {SSIM}{$\uparrow$} / {LPIPS}{$\downarrow$} \\
       \midrule
       {Bicubic Degradation Pre-training}  & {28.08 / 0.8357 / 0.397}  \\ 
       {SRA Fine-tuning~\cite{DCSR}}  & {28.11 / 0.8109 / 0.268}  \\ 
       {Our Fine-tuning}  & {29.14 / 0.8511 / 0.185}  \\ 
      \bottomrule
    \end{tabular}
    \end{center}
 \vspace{-2mm}
\end{table}

\begin{table}[t] 
  \small
  \caption{Ablation study on two-stage alignment methods. } 
  \label{tab:ablation_alignment}
  \centering\noindent
  \centering%
 \vspace{-4mm}
  \begin{center}
    \begin{tabular}{ccc}
      \toprule  
      \tabincell{c}{Patch-based \\ Optical Flow} & 
      \tabincell{c}{Auxiliary-LR \\ Guiding}  & 
      {PSNR}{$\uparrow$} / {SSIM}{$\uparrow$} / {LPIPS}{$\downarrow$}  \\
        \midrule
       \texttimes  & \texttimes & 29.19 / 0.8495 / 0.199 \\ 
       \checkmark  & \texttimes    & 29.52 / 0.8533 / 0.186 \\ 
       \checkmark  & \checkmark  & 29.62 / 0.8582 / 0.187 \\ 
      \bottomrule
    \end{tabular}
    \end{center}
 \vspace{-2mm}
\end{table}

\begin{table}[t!] 
  \small
  \caption{Ablation study on loss terms of auxiliary-LR generator. $\lambda_\mathit{p}$ denotes the coefficient of position preserving loss in Eqn.~(\ref{eqn:loss_d}).}
  \label{tab:auxiliary_loss}
  \centering\noindent
  \centering%
 \vspace{-4mm}
  \begin{center}
    \begin{tabular}{ccccc}
      \toprule  
      {$\lambda_\mathit{p}$} &  0 & 1 & 100 & 10000 \\
      \midrule
       PSNR$\uparrow$ & 29.37 & 29.43  & 29.62  & 29.61  \\
      \bottomrule
    \end{tabular}
    \end{center}
\end{table}

\begin{table}[t!] 
  \small
  \vspace{-2mm}
  \caption{Ablation study on the noise of auxiliary-LR.}
  \vspace{-1mm}
  \label{tab:ablation_noise}
  \centering\noindent
  \centering%
 \vspace{-3mm}
  \begin{center}
    \begin{tabular}{cc}
      \toprule  
      Noise & 
      {PSNR}{$\uparrow$} / {SSIM}{$\uparrow$} / {LPIPS}{$\downarrow$} \\
       \midrule
       None  &  29.39 / 0.8550 /  0.233 \\ 
       JPEG  &  29.26 / 0.8523 / 0.198 \\
       Gaussian &  29.61 / 0.8560 / 0.187 \\
       Gaussian and JPEG  & 29.62 / 0.8582 / 0.187 \\   
      \bottomrule
    \end{tabular}
    \end{center}
 \vspace{-2mm}
\end{table}

\begin{table}[t!] 
  \small
  \setlength{\tabcolsep}{1mm}
  \caption{Ablation study on AdaSTN.}
  \label{tab:ablation_AdaSTN}
  \centering%
  \vspace{-4mm}
  \begin{center}
    \begin{tabular}{lcc}
      \toprule
      Method &  {PSNR}{$\uparrow$} / {SSIM}{$\uparrow$} / {LPIPS}{$\downarrow$}  \\
       \midrule
       Baseline & 29.52 / 0.8533 / 0.186 \\ 
       Baseline + Deformable Conv~\cite{DefConv}  & 29.60 / 0.8582 / 0.191 \\ 
       Baseline + AdaSTN & 29.62 / 0.8582 / 0.187 \\ 
      \bottomrule
    \end{tabular}
    \end{center}
  \vspace{-3mm}
\end{table}

\subsection{Effect of Alignment between LR and GT}
\noindent\textbf{Effect of Two-stage Alignment.}
  In order to evaluate the effect of our two-stage alignment method (see Sec.~\ref{sec:alignment_lr}), we first remove patch-based optical flow alignment and auxiliary-LR guiding alignment to train a baseline model in our self-supervised manner.
  Then we add the alignment method of these two stages in turn to experiment.
  When taking patch-based optical flow alignment only, the PSNR increases by 0.33 dB against the baseline, as shown in Table~\ref{tab:ablation_alignment}.
  Coupled with auxiliary-LR guiding alignment, better quantitative results can be further attained.

\noindent\textbf{Effect of Auxiliary-LR Generator.}
  We show the auxiliary-LR images before adding synthetic noise $\mathbf{n}$ in Fig.~\ref{fig:misalignment}(b).
  And the corresponding LR and GT images are shown in Fig.~\ref{fig:misalignment}(a) and Fig.~\ref{fig:misalignment}(c), respectively.
  The red lines and arrows in the same row are in the same position relative to the image.
  It can be seen that the auxiliary-LR has similar contents as LR and is aligned with GT.
  It indicates that the function of the auxiliary-LR generator is guaranteed.
  In addition, we conduct an experiment that adds noise $\mathbf{n}$ to bicubic downsampling GT and replaces auxiliary-LR with it. In this case, PSNR drops by 0.88 dB, and LPIPS gets worse by 0.078.
  The result shows the auxiliary-LR generator is necessary and effective.
  We also conduct experiments on different coefficients (\ie, $\lambda_\mathit{p}$) of position preserving loss, as shown in Table~\ref{tab:auxiliary_loss}.
  In order to bring auxiliary-LR into play better on alignment and obtain better SR performance, we take a trade-off between content preserving loss and position preserving loss, and set $\lambda_\mathit{p}$ to 100.

\noindent\changes
{\textbf{Effect of Noise in Auxiliary-LR.}
Some noise is added to auxiliary-LR to prevent overfitting problems of the restoration module.
Gaussian noise is a natural choice, and we empirically find it is sufficient to achieve the goal, as shown in Table~\ref{tab:ablation_noise}.
Moreover, additional JPEG compression noise can provide further slight improvement, which can be regarded as simulated artifacts.
}

\noindent\textbf{Effect of AdaSTN.}
  We regard the model only using patch-based optical flow alignment as the baseline.
  And we modify AdaSTN to deformable convolution~\cite{DefConv} to verify its effect.
  Specifically, instead of calculating the offset by estimating the affine transformation matrix and vector according to Eqn.~(\ref{eqn:offset}), we directly estimate the offset for deformable convolution.
  The results in Table~\ref{tab:ablation_AdaSTN} indicate that AdaSTN has superior performance than deformable convolution.

\begin{table}[t!] 
  \small
  \centering
  \setlength{\tabcolsep}{1mm}
  \caption{Quantitative results comparison while using different loss terms.}
  \label{tab:loss}
  \centering\noindent
  \centering%
  \vspace{-3mm}
  \begin{center}
    \begin{tabular}{lc}
      \toprule
      Loss Terms &
      {PSNR}{$\uparrow$} / {SSIM}{$\uparrow$} / {LPIPS}{$\downarrow$} \\
        \midrule
        $\ell_1$  & 29.74 / 0.8708 / 0.280 \\ 
        $\ell_1$ + Perceptual + Adversarial~\cite{RelateGAN}   & 29.27 / 0.8463 / 0.213 \\
        $\ell_1$ + SW~\cite{SelfDZSR}  & 29.20 / 0.8479 / 0.188 \\
        $\ell_1$ + LOSW  & 29.62 / 0.8582 / 0.187 \\ 
      \bottomrule  
    \end{tabular}
    \end{center}
    \vspace{-4mm}
\end{table}

\begin{figure*}[t]
	\centering
	\subfloat[]
	{
		\begin{minipage}{0.15\linewidth}
			\centering
			\includegraphics[width=\linewidth]{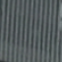}
		\end{minipage}
	}
	\hspace{-0.1cm}
	\subfloat[] 
	{
		\begin{minipage}{0.15\linewidth}
			\centering
			\includegraphics[width=1\linewidth]{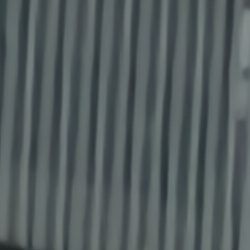}
		\end{minipage}
	}
	\hspace{-0.1cm}
	\subfloat[] 
	{
		\begin{minipage}{0.15\linewidth}
			\centering
			\includegraphics[width=\linewidth]{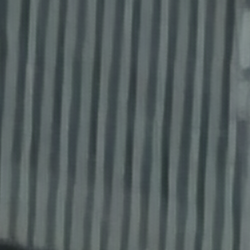}
		\end{minipage}
	}
	\hspace{-0.1cm}
	\subfloat[] 
	{
		\begin{minipage}{0.15\linewidth}
			\centering
			\includegraphics[width=\linewidth]{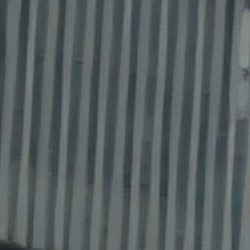}
		\end{minipage}
	}
	\hspace{-0.1cm}
	\subfloat[] 
	{
		\begin{minipage}{0.15\linewidth}
			\centering
			\includegraphics[width=\linewidth]{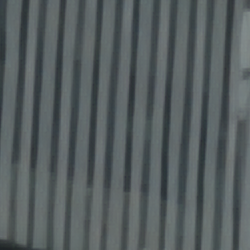}
		\end{minipage}
	}
	\hspace{-0.1cm}
	\subfloat[]
	{
		\begin{minipage}{0.15\linewidth}
			\centering
			\includegraphics[width=\linewidth]{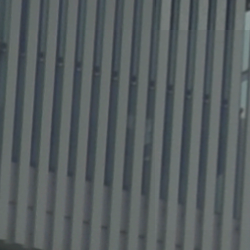}
		\end{minipage}
	}%
        \vspace{-6mm}
        
 	\subfloat[LR]
	{
		\begin{minipage}{0.15\linewidth}
			\centering
			\includegraphics[width=\linewidth]{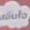}
		\end{minipage}
	}
	\hspace{-0.1cm}
	\subfloat[$\ell_1$ loss] 
	{
		\begin{minipage}{0.15\linewidth}
			\centering
			\includegraphics[width=1\linewidth]{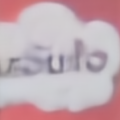}
		\end{minipage}
	}
	\hspace{-0.1cm}
	\subfloat[$\ell_1$+Per.+Adv. loss~\cite{RelateGAN}] 
	{
		\begin{minipage}{0.15\linewidth}
			\centering
			\includegraphics[width=\linewidth]{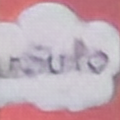}
		\end{minipage}
	}
	\hspace{-0.1cm}
	\subfloat[$\ell_1$ + SW loss~\cite{SelfDZSR}] 
	{
		\begin{minipage}{0.15\linewidth}
			\centering
			\includegraphics[width=\linewidth]{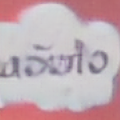}
		\end{minipage}
	}
	\hspace{-0.1cm}
	\subfloat[$\ell_1$ + LOSW loss (Ours)] 
	{
		\begin{minipage}{0.15\linewidth}
			\centering
			\includegraphics[width=\linewidth]{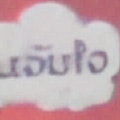}
		\end{minipage}
	}
	\hspace{-0.1cm}
	\subfloat[GT]
	{
		\begin{minipage}{0.15\linewidth}
			\centering
			\includegraphics[width=\linewidth]{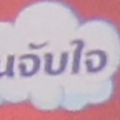}
		\end{minipage}
	}%
  \vspace{-2mm}
  \caption{Visual results comparison when using different loss terms. `Per.' and `Adv.' represent perceptual and adversarial loss terms, respectively. The textures and details in our results are more realistic.}
  \label{fig:ablation_loss}
  \vspace{-2mm}
\end{figure*}

\subsection{Effect of LOSW Loss}
  Most RefSR methods~\cite{SRNTT,TTSR,MASA-SR,C2-Matching} adopt VGG-based~\cite{VGGLOSS} perceptual loss and adversarial loss~\cite{GAN} for more realistic results.
  Here we follow $C^2$-Matching~\cite{C2-Matching} to train SelfTZSR++ using a combination of $\ell_1$ reconstruction loss, perceptual loss, and adversarial loss based on Relativistic GAN~\cite{RelateGAN}.
  The quantitative results are shown in Table~\ref{tab:loss}.
  It can be seen that the model trained by LOSW loss obtains a 0.35 dB PSNR and 0.026 LPIPS gain than that by adversarial loss.
  We also take SW~\cite{SelfDZSR} loss to train a model for comparison.
  The model trained by LOSW loss also has a higher PSNR metric than that by SW loss, while the gap of the LPIPS metric is small.
  In comparison with the model only using $\ell_1$ loss, the PSNR of the model using LOSW loss is only 0.12 dB worse, while LPIPS has an advantage of 0.093.
  Unlike other loss terms that drop fidelity metrics when obtaining better perceptual metrics, LOSW loss has a superior ability to measure perceptual differences while maintaining fidelity.
  
  Fig.~\ref{fig:ablation_loss} shows a visual result comparison when using different loss terms.
  When only $\ell_1$ loss is taken, the result is over-smooth.
  Although adversarial loss and SW loss bring sharper content, they lead to some unrealistic artifacts.
  LOSW can help generate more satisfactory images which are more faithful to the high-resolution ground-truth and has fewer artifacts.
  In short, LOSW loss can achieve a better trade-off in fidelity and perception.

\subsection{Effect of Different Refs and Fusion Schemes}

\noindent\textbf{Effect of Different Refs.}
We conduct experiments with different reference images ($\times r_w$ Ref from ultra-wide image and $\times r_t$ Ref from telephoto image).
As shown in Table~\ref{tab:diff_ref}, using $\times r_t$ Ref leads to higher PSNR compared to using $\times r_w$ Ref. And it further improves fidelity and perceptual metrics when combining both Refs.

\noindent\textbf{Effect of Refs Fusion Scheme.}
Using both reference images, we investigate various fusion schemes between aligned Ref features and aligned LR ones in our experiments.
From Table~\ref{tab:fusion}, it can not achieve satisfactory results when directly concatenating features from Refs and LR image together for restoration.
The proposed progressive fusion scheme first concatenates $\times r_w$ Ref features and the LR ones, and processes them. Then, the processed features are fused with $\times r_t$ Ref features for further modulation.
It can achieve a 0.49 dB PSNR improvement over the strategy of directly concatenating.
In addition, we also conduct an experiment by reversing the fusion order of the two Ref features, showing a 0.22 dB PSNR drop over the proposed scheme.
The experiment illustrates that it is more suitable to fuse LR features first with lower resolution ($\times r_w$) Ref features, and then with higher resolution ($\times r_t$) Ref ones.

\begin{table}[t] 
  \small
  \centering
  \caption{Quantitative results comparison while using different Refs.}
  \label{tab:diff_ref}
  \centering\noindent
  \centering%
  \vspace{-3mm}
  \begin{center}
    \begin{tabular}{cc}
      \toprule
      Ref Images &
      {PSNR}{$\uparrow$} / {SSIM}{$\uparrow$} / {LPIPS}{$\downarrow$} \\
        \midrule
       $\times r_w$ Ref   & 29.04 / 0.8481 / 0.201 \\ 
       $\times r_t$ Ref  & 29.30 / 0.8511 / 0.201 \\
       $\times r_w$ Ref and $\times r_t$ Ref  & 29.62 / 0.8582 / 0.187 \\ 
      \bottomrule  
    \end{tabular}
    \end{center}
    \vspace{-1mm}
\end{table}

\begin{table}[t] 
  \small
  \centering
  \caption{Quantitative results of different fusion schemes between aligned Ref features and aligned LR ones.}
  \label{tab:fusion}
  \centering\noindent
  \centering%
  \vspace{-3mm}
  \begin{center}
    \begin{tabular}{cc}
      \toprule
      Fusion Schemes &
      {PSNR}{$\uparrow$} / {SSIM}{$\uparrow$} / {LPIPS}{$\downarrow$} \\
        \midrule
        Concat  $\times r_w$ Ref and $\times r_t$ Ref & 29.13 / 0.8517 / 0.186  \\ 
       $\times r_t$ Ref firstly, then $\times r_w$ Ref & 29.40 / 0.8572 / 0.189  \\
       $\times r_w$ Ref firstly, then $\times r_t$ Ref &  29.62 / 0.8582 / 0.187 \\ 
      \bottomrule  
    \end{tabular}
    \end{center}
    \vspace{-3mm}
\end{table}

\begin{table*}[t!] 
  \small
  \caption{Effect of scaling up models. The models are trained with their all loss terms. The PNSR, SSIM, and LPIPS metrics are calculated on the full image. The \#FLOPs is measured when $\times4$ super-resolving LR image to  $1280\times720$ resolution.}
  \label{tab:large-all-loss}
  \centering\noindent
  \centering%
  \vspace{-4mm}
  \begin{center}
  \scalebox{1}{
    \begin{tabular}{lcccc}
      \toprule
      \multirow{2}{*}{Method} & \multirow{2}{*}{\# Params (M)} & \multirow{2}{*}{\#FLOPs (G)} 
      & \multicolumn{2}{c}{{PSNR}{$\uparrow$} {/} {SSIM}{$\uparrow$} {/} {LPIPS}{$\downarrow$}} \\
      & & &  \textbf {Nikon Camera Images}
      & \textbf {Iphone Camera Images} \\ 
       \midrule
         SelfDZSR++  & 3.1 & 454     & 29.30 / 0.8511 / 0.201 & 23.64 / 0.7266 / 0.244   \\ 
         SelfTZSR++  & 3.3 & 538  & 29.62 / 0.8582 / 0.187  & 24.00 / 0.7466 / 0.215  \\  
         \midrule
         SelfDZSR++ (Large)  & 16.9 & 1981    & 29.42 / 0.8508 / 0.186   & 23.49 / 0.7191 / 0.232     \\ 
         SelfTZSR++  (Large)   & 17.2 & 2084   & 29.72 / 0.8594 / 0.173   & 24.13 / 0.7525 / 0.204    \\  
      \bottomrule
    \end{tabular}
    }
    \end{center}
    \vspace{-3mm}
\end{table*}

\subsection{\changes{Effect of Scaling up Models}}
\changes{
Here we scale up our models to conduct experiments.
To achieve a comparable computational cost with competitive SISR methods, we double the number of channels and triple the depth in the restoration module, named `SelfDZSR++ (Large)' and `SelfTZSR++ (Large)'.
The quantitative results are shown in Table~\ref{tab:large-all-loss}.
It can be seen that scaling up models generally brings about performance improvements, especially on the LPIPS metric.
}

\section{Conclusion}
  Real-world image super-resolution from dual zoomed observations (DZSR) is an emerging topic, which aims to super-resolve the ultra-wide image with the reference of telephoto image.
  To circumvent the problem that ground-truth is unavailable, we propose an effective self-supervised learning method.
  To mitigate the adverse effect of image misalignment during training, we propose a two-stage alignment method consisting of patch-based optical flow and auxiliary-LR guiding alignment.
  To obtain visually pleasing results, we present local overlapped sliced Wasserstein loss.
  Moreover, we extend DZSR to multiple zoomed observations,  where we present a progressive fusion scheme for better restoration.
  Experiments show that our proposed method can achieve better performance against the state-of-the-art methods both quantitatively and qualitatively.

\section*{Acknowledgment}
  This work was supported by the National Natural Science Foundation of China (NSFC) under Grants No.s 61872118 and U19A2073.

\bibliographystyle{IEEEtran}
\bibliography{IEEEabrv,egbib}


\end{document}